\def\cvprPaperID{6420}
\def\confName{ICCV}
\def\confYear{2023}

\def\authorBlock{
    \quad\quad Daiqing Li $^{1^*}$ \quad\quad Huan Ling$^{1,2,3^*}$  
    \quad\quad Amlan Kar$^{1,2,3}$ \quad\quad David Acuna$^{1,2,3}$
\and
\quad\quad Seung Wook Kim$^{1,2,3}$ 
  \quad Karsten Kreis$^{1}$ \quad Antonio Torralba$^{4}$ \quad Sanja Fidler$^{1,2,3}$  \vspace{2mm}\\
  
  \small{\textsuperscript{1}NVIDIA \quad \textsuperscript{2}University of Toronto \quad \textsuperscript{3}Vector Institute \quad \textsuperscript{4}MIT \vspace{1pt}}\\
  \scriptsize \textit{Project page:} \url{https://research.nvidia.com/labs/toronto-ai/DreamTeacher/} \vspace{2mm}\\ 
}

\newif\ifreview 
\newif\ifarxiv \newcommand{\arxiv}{\arxivtrue}
\newif\ifcamera 
\newif\ifrebuttal 

\arxiv 

\pdfoutput=1
\documentclass[10pt,twocolumn,letterpaper]{article}
\ifreview \usepackage[review]{cvpr} \fi
\ifarxiv \usepackage[pagenumbers]{cvpr} \fi
\ifrebuttal \usepackage[rebuttal]{cvpr} \fi
\ifcamera \usepackage{cvpr} \fi

\usepackage{graphicx}
\usepackage{amsmath}
\usepackage{amssymb}
\usepackage{booktabs}

\usepackage{times}
\usepackage{microtype}
\usepackage{epsfig}
\usepackage[table,xcdraw]{xcolor}
\usepackage{caption}
\usepackage{float}
\usepackage{placeins}
\usepackage{color, colortbl}
\usepackage{stfloats}
\usepackage{enumitem}
\usepackage{tabularx}
\usepackage{xstring}
\usepackage{multirow}
\usepackage{xspace}
\usepackage{url}
\usepackage{subcaption}
\usepackage{xcolor}
\usepackage[hang,flushmargin]{footmisc}
\usepackage{amsmath}
\usepackage{amssymb}
\ifcamera \usepackage[accsupp]{axessibility} \fi

% ...

\newcommand{\nbf}[1]{{\noindent \textbf{#1.}}}

\ifarxiv  \fi

\newcommand*\rot{\rotatebox{90}}
\newcommand{\R}[1]{{%
    \textbf{%
        \ifstrequal{#1}{1}{\textcolor{red}{R#1}}{%
        \ifstrequal{#1}{2}{\textcolor{blue}{R#1}}{%
        \ifstrequal{#1}{3}{\textcolor{magenta}{R#1}}{%
        \ifstrequal{#1}{4}{\textcolor{teal}{R#1}}{%
                           \textcolor{cyan}{R#1}%
        }}}}%
    }%
}}

\usepackage{xr-hyper}

\makeatletter
\newcommand*{\addFileDependency}[1]{
  \typeout{(#1)}
  \@addtofilelist{#1}
  \IfFileExists{#1}{}{\typeout{No file #1.}}
}

\makeatother

\usepackage[pagebackref,breaklinks,colorlinks]{hyperref}
\usepackage[capitalize]{cleveref}
\crefname{section}{Sec.}{Secs.}
\crefname{table}{Table}{Tables}
\crefname{figure}{Fig.}{Figs.}

\newcommand{\ourmodel}{DreamTeacher}

\definecolor{LightGreen}{rgb}{0.9,1,0.9}
\definecolor{LightGray}{rgb}{0.95,0.95,0.95}
\definecolor{LightBlue}{rgb}{0.8,0.9,1.0}
\begin{document}
\title{DreamTeacher: Pretraining Image Backbones with Deep Generative Models}
\author{\authorBlock}
\newcommand\blfootnote[1]{%
  \begingroup
  \renewcommand\thefootnote{}\footnote{#1}%
  \addtocounter{footnote}{-1}%
  \endgroup
}

\twocolumn[{
    \renewcommand\twocolumn[1][]{#1}
    \maketitle
    \begin{center}

    \vspace{-23pt}
    \centering\

    \begin{minipage}{0.95\linewidth}
        \centering
       \vspace{-2mm}
    \includegraphics[width=\linewidth]{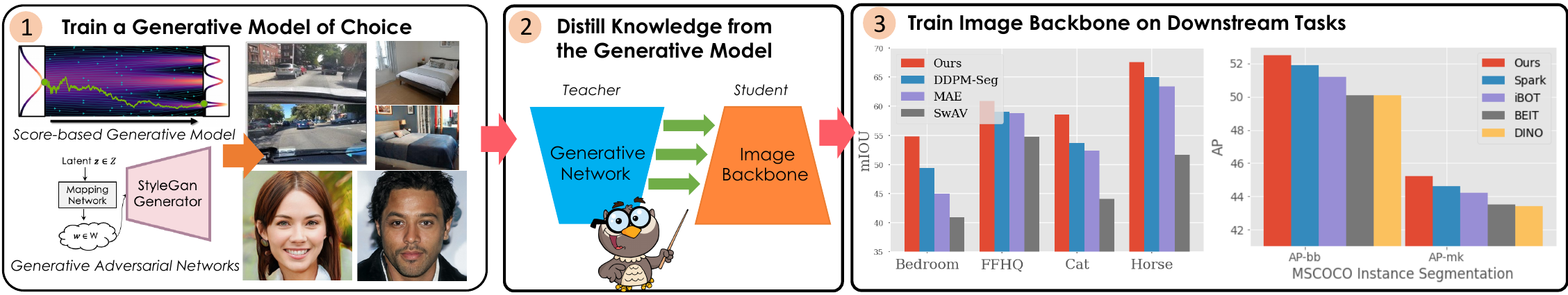}
    \end{minipage}
    \vspace{-2.8mm}
    \captionof{figure}{
        \small We propose {\ourmodel}, a  framework for distilling knowledge from a pre-trained generative network onto a target image backbone, as a generic pre-training mechanism that doesn't require labels. We investigate \emph{feature distillation}, and optionally \emph{label distillation} (when task-specific labels are available). Our {\ourmodel} outperforms existing self-supervised methods on a variety of benchmarks.  
    }
    \label{fig:teaser}
    \vspace{3pt}

    \end{center}
}]

\blfootnote{$^\ast$ Equal Contribution.}

\vspace{-6pt}
\begin{abstract}
\vspace{-4pt}

In this work, we introduce a self-supervised feature representation learning framework {\ourmodel} that utilizes generative networks for pre-training downstream image backbones. 
We propose to distill knowledge from a trained generative model into standard image backbones that have been well engineered for specific perception tasks.
We investigate two types of knowledge distillation: 1) distilling learned {\emph generative features} onto target image backbones as an alternative to pretraining these backbones on large labeled datasets such as ImageNet, and 2) distilling \emph{labels} obtained from generative networks with task heads onto logits of target backbones. 
We perform extensive analyses on multiple generative models, dense prediction benchmarks, and several pre-training regimes. 
We empirically find that our {\ourmodel} significantly outperforms existing self-supervised representation learning approaches across the board. 
Unsupervised ImageNet pre-training with DreamTeacher leads to significant improvements over ImageNet classification pre-training on downstream datasets, showcasing generative models, and diffusion generative models specifically, as a promising approach to representation learning on large, diverse datasets without requiring manual annotation. 
\end{abstract}
\vspace{-5.5mm}
\section{Introduction}
\label{sec:intro}
\vspace{-0.5mm}

Self-supervised representation learning is becoming an effective way of pre-training vision backbones~\cite{chen2020simple,he2020momentum, grill2020bootstrap, caron2020unsupervised, chen2020improved}. The premise of this line of work is to leverage large unlabeled datasets as additional source of training data in order to boost performance of downstream networks, and to reduce the need for large labeled target datasets. Recent works have shown that self-supervised pre-training on ImageNet can now come close to supervised pre-training, even outperforming it on some downstream datasets and tasks such as pixelwise semantic and instance segmentation~\cite{he2020momentum, chen2020improved, wang2021dense}. 

One of the dominant approaches to self-supervised representation learning are variants of contrastive learning, where the target backbone is trained to map transformed views of an image closer in latent space than images randomly drawn from the dataset~\cite{chen2020simple}. Improvements to this paradigm include introducing spatial losses~\cite{wang2021dense, xiong2020loco, xie2021propagate, xie2021detco}, and improving training stability 
with fewer or no
negative examples~\cite{he2020momentum, chen2020improved,grill2020bootstrap, chen2021exploring}. 

Another line of work pursues reconstruction losses for supervision, where certain regions get masked from an input image, and backbones get trained to reconstruct them~\cite{doersch2015unsupervised, he2022masked, xie2022simmim, wei2022masked}, also known as Masked Image Modeling (MIM). 
This task is mostly treated as deterministic, ie supervising a single explanation for the masked region.
This line of work typically investigates masking strategies, architecture design and training recipes to train better backbones. These methods have achieved state-of-the-art (SoTA) performance when applied to Vision Transformer-based backbones; however, recently sparse CNN-based image backbones~\cite{tian2023designing} have been shown to be as performant.

In this paper, we argue for generative models as representation learners: for the simplicity of the objective -- to generate data, and intuitive representational power -- generating high quality samples as an indication of learning semantically capable internal representations. Using generative networks as representation learners is not a novel concept. DatasetGAN and variants~\cite{zhang2021datasetgan, li2022bigdatasetgan, baranchuk2021labelefficient} proposed to add task-dependent heads on top of StyleGAN's or a diffusion model's features, and used these augmented networks as generators of labeled data, on which downstream networks are then trained.  SemanticGAN~\cite{li2021semantic} instead used StyleGAN with an additional  task decoder as the task network itself -- by encoding images into the latent space of the generative model and using the task head for producing perception output.\looseness=-1

We introduce {\ourmodel}, a representation learning framework  that leverages generative models for pre-training downstream perception models via distillation.
We investigate two types of distillation: 1) feature distillation, where we propose methods for distilling generative features to target backbones, as a general pre-training mechanism that does not require any labels. 2) label distillation: using task-heads on top of generative networks for distilling knowledge from a labeled dataset onto target backbones, in a semi-supervised regime. We focus our work on diffusion models~\cite{sohl2015deep,ho2020denoising,song2020score} and GANs~\cite{goodfellow2014generative,karras2019style,karras2020analyzing} as the choice of generative models. 
For target backbones, we focus on CNNs, for two major reasons. 1) CNN-based backbones have been shown to achieve SoTA representation learning performance for both contrastive and MIM approaches~\cite{liu2022convnet, tian2023designing, Woo2023ConvNeXtV2, wang2022internimage}, 2) SoTA generative models today (GANs and diffusion models) primarily still use CNNs internally. In preliminary experiments, we also explored vision transformer backbones, but found it challenging to distill features from CNN-based generative models into vision transformers. Generative models built with vision transformer architectures are nascent~\cite{bao2022all, Peebles2022DiT}, and hence we leave a thorough exploration of {\ourmodel} with these architectures to future work.\looseness=-1

We experimentally show that {\ourmodel} outperforms existing self-supervised learning approaches on various benchmarks and settings.  
Most notably, when pre-trained on ImageNet without any labels, our method 
significantly outperforms methods that are pre-trained on ImageNet with full supervision, on several dense prediction benchmarks and tasks such as semantic segmentation on ADE20K~\cite{zhou2019semantic}, instance segmentation on MSCOCO~\cite{lin2014microsoft} and on the autonomous driving dataset BDD100K~\cite{yu2020bdd100k}. 
On object-focused datasets with millions of unlabeled images~\cite{zhang2021datasetgan,yu2015lsun}, our method, when trained solely on the target domain, significantly outperforms variants that are pre-trained on ImageNet with label supervision, and achieves new SoTA results. These results highlight generative models, especially diffusion-based generative models~\cite{song2020score, ho2020denoising, dhariwal2021diffusion}, as powerful representation learners that can effectively leverage diverse unlabeled datasets at scale. 
\section{Related Work}
\label{sec:related}

\begin{figure*}[h!]
\vspace{-2mm}
\begin{center}
\includegraphics[width=0.87\linewidth,trim=0 10 0 0,clip]{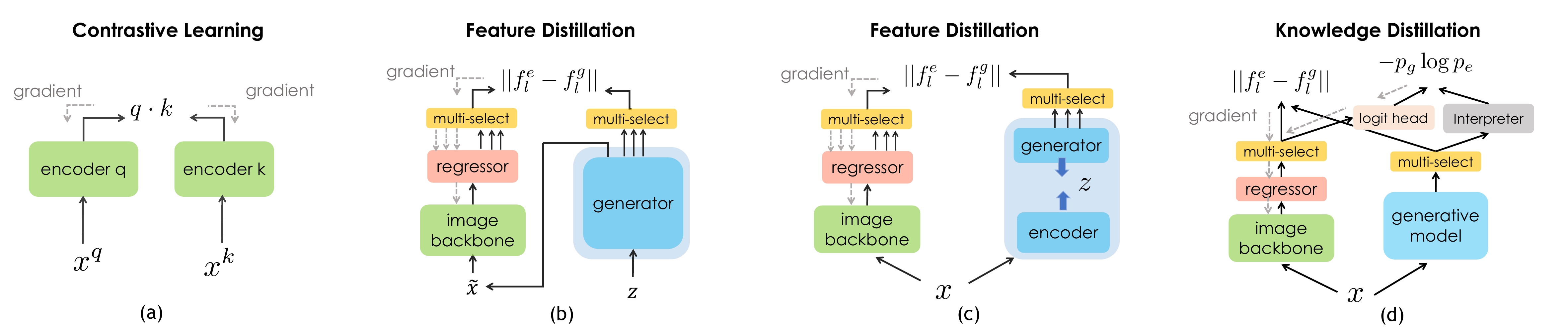}
\end{center}
\vspace{-6mm}
\caption{
Different representation learning approaches: {\bf (a)} a representative discriminative pretraining using a siamese-based network and contrastive loss, {\bf (b)} our {\ourmodel} generative pretraining framework when sampling examples from the generative model, {\bf (c)} our  {\ourmodel} generative pretraining framework on encoded real data, {\bf (d)} our mix distillation when a small number of labels are available (20-40 labeled data in our experiments). Multi-select means selecting features from different layers.
}
\label{fig:overview}
\vspace{-3mm}
\end{figure*}

\nbf{Discriminative Representation Learning}
Early 
representation learning methods relied on handcrafted pretext tasks such as relative patch prediction \cite{doersch2015unsupervised}, solving jigsaw puzzles \cite{noroozi2016unsupervised}, colorization~\cite{zhang2016colorful}, and relative rotation~\cite{gidaris2018unsupervised}. 
Instead, our pretext task is to predict features of a pretrained generative model, which in turn is trained with a simple and natural objective: maximize the log likelihood of the image data. 
The ability to synthesize and manipulate high quality samples 
is promising sign that generative networks learn both semantic and geometric knowledge internally~\cite{zhang2021datasetgan}.

Recent breakthroughs come from contrastive representation learning methods~\cite{chen2020simple, chen2020improved, grill2020bootstrap}. SimCLR\cite{chen2020simple} was the first to show competitive results in linear probing and transfer learning without using class labels, compared to supervised pre-training. Follow-up works MoCo~\cite{he2020momentum}, MoCoV2~\cite{chen2020improved} and BYOL\cite{grill2020bootstrap} improve over the siamese network design with a memory bank and gradient stopping. However, these methods rely on heavy data augmentation~\cite{xiao2020should} and heuristics to select the negative examples. This may not generalize well to datasets beyond well-curated object-centric datasets like ImagetNet~\cite{51319}.

Another line of work~\cite{wang2021dense,xiong2020loco,xie2021propagate,henaff2021efficient}  aims to improve over the global contrastive objective and focuses on region-based features which are useful for dense prediction tasks. denseCL~\cite{wang2021dense} extends MoCoV2~\cite{chen2020improved} to predict auxiliary dense features, PixPro~\cite{xie2021propagate} extends BYOL~\cite{grill2020bootstrap} to have pixel-wise consistency across two views, while DetCon~\cite{henaff2021efficient} introduces masked pooling to focus on object-wise features. However, these methods require special designs for certain tasks~\cite{xie2021propagate,xie2021detco}, or  additional heuristics 
for complex scene datasets~\cite{henaff2021efficient}. In our work, we focus on generative networks for representation learning specifically focused on various dense prediction tasks. 

\nbf{Generative Representation Learning}
The ideas of leveraging generative models for learning representations for recognition tasks dates back to Hinton~\cite{hinton07}. 
Recent works use advanced generative models and techniques to develop representation learning methods. BiGAN~\cite{donahue2016adversarial} proposed to jointly train an encoder with adversarial training objective.
BigBiGAN~\cite{donahue2019large} leveraged the advancement of BigGAN~\cite{brock2018large} and showed competitive linear probing results in ImageNet. Methods like iGPT~\cite{chen2020generative} and VIM~\cite{yu2021vector} pre-train large transformer networks with autoregressive generative pre-training objectives 
, achieving compelling linear probing results on ImagetNet, but they did not show results on dense prediction tasks. 
Furthermore, these methods train a single image backbone with both discriminative and generative objectives and thus cannot leverage the specific designs for each.  

DatasetGAN~\cite{zhang2021datasetgan,li2022bigdatasetgan} was among the first to show that a pretrained GAN can significantly benefit perception tasks, especially in the low labeled data regime. Specifically, the authors  added a task-specific head on top of StyleGAN and  synthesized a labeled dataset for training downstream perception networks. SemanticGAN~\cite{li2021semantic} proposed to model the joint distribution of images and labels. Inference was performed by first encoding the test images into the latent space of StyleGAN and then decoded the labels using the task-head. DDPM-seg~\cite{baranchuk2021labelefficient} followed this line of work but used a denoising diffusion probabilistic model (DDPMs) instead of StyleGAN. 
In our paper, we continue this line of work but focus on distilling knowledge from a pre-trained generative model, diffusion model specifically, to downstream image backbones as a general way of pre-training. We provide an extensive evaluation of generative networks in the context of representation learning on various benchmarks and tasks. 


\nbf{Knowledge Distillation} Hinton et al~\cite{hinton2015distilling} were first to propose knowledge distillation as an effective means of improving performance -- with the idea of  distilling logits from a large teacher network into a smaller student network. FitNets~\cite{romero2014fitnets} proposed to mimick the teacher's intermediate feature activations as additional hints for the student network. Follow-up works try to utilize different forms of knowledge from the teacher network:  spatially~\cite{zagoruyko2016paying}, channel-wisely~\cite{shu2021channel}, and from multi-levels~\cite{chen2021distilling}. Usually, the teacher and student networks share a similar training objective, the network architecture, and require labels to train the teacher network. In our work, our generative model is treated as a teacher, and is trained without labels and the objective is not task-specific. Our student networks are image backbones of choice, which might not share a similar architecture as the teacher.

\begin{figure*}[h]
\begin{center}
\vspace{-4mm}
\includegraphics[width=0.85\linewidth]{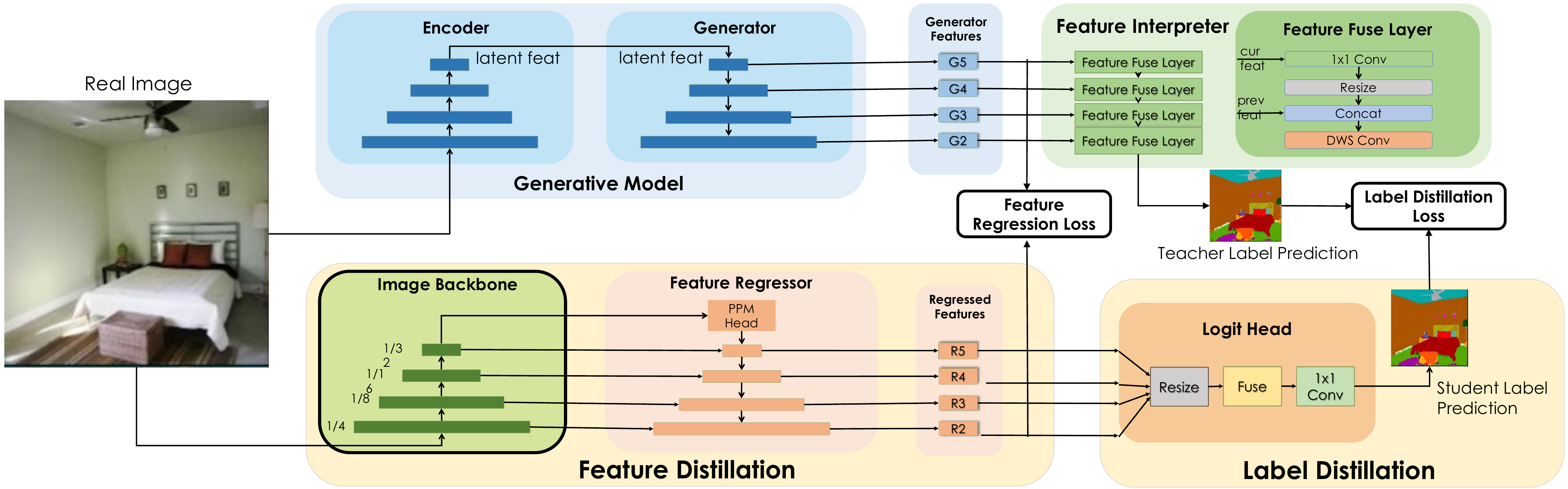}
\end{center}
\vspace{-5mm}
\caption{
\textbf{\footnotesize{\ourmodel} architecture}: Feature regression module (FR) maps and fuses multi-scale features of a (CNN) image backbone. We supervise FR with features from the generator's decoding network. We optionally add a \emph{feature interpreter}~\cite{zhang2021datasetgan} to the generator to train a  task head with supervised labels -- used to supervise the image backbone with label distillation loss. 
}
\label{fig:framework}
\vspace{-4mm}
\end{figure*}

\vspace{-2mm}
\section{{\ourmodel} Framework}
\label{sec:method}
We describe our {\ourmodel} framework in the context of two scenarios: unsupervised representation learning where no labels are available during pre-training, and semi-supervised learning where a fraction of labels are available. 

We utilize a trained generative model $G$ and \emph{distill} its learned representation into a target image backbone $f$. 
Our recipe for training $f$ remains the same in both scenarios and choices of $G$ and $f$. First, we create a \emph{feature dataset} $D=\{x_i, \mathbf{f}^g_i\}_{i=1}^{N}$ of images $x_i$ and corresponding features $\bf{f}^g_i$ extracted from the generative model. Next, we train $f$ using the dataset $D$ by distilling features $\mathbf{f}^g_i$ into the intermediate features of $f(x_i)$.  We focus on convolutional backbones $f$, leaving exploration into transformers for future work. We drop subscript $i$ for brevity from here on.

In Sec.~\ref{sec:unsupreps}, we describe the design of our unsupervised distillation process. We tackle the semi supervised regime in Sec.~\ref{sec:labelreps}, where labels are available on a fraction of the pre-training dataset. 
\vspace{-2mm}
\subsection{Unsupervised Representation Learning}
\label{sec:unsupreps}

For unsupervised representation learning given a feature dataset $D$, we attach feature regressors at different hierarchical levels of the backbone $f$ to regress the corresponding generative features $\mathbf{f}^g_i$ from an image $x_i$. We first discuss creating a feature dataset, followed by the design of feature regressors and end by introducing our distillation objective.

\nbf{Creating a feature dataset $D$} Generative models give us two distinct ways of creating our desired feature dataset $D$. One could sample images from the generative model $G$ and record intermediate features from the generative process. In principle, this could synthesize datasets of infinite size, but may suffer from issues such as mode dropping, where the generative model may not have learned some parts of the distribution sufficiently well. We refer to such a dataset as a \emph{synthesized dataset}. Instead, one could encode real images, labeled or unlabeled, into the latent space of the generative model $G$, using an encoding process. We refer to such a dataset as an \emph{encoded dataset}.

A \emph{synthesized dataset} $D$ is created by sampling images $\tilde{x} \sim G(z)$, where $z$ is sampled from the generative model $G$'s prior distribution. We record hierarchical intermediate features from $G(z)$ as $\mathbf{f}^g = \{f^g_l\}_{l=1}^L$ where $l$ denotes the hierarchy level of the features from a total of $L$ levels. 
We employ this approach when using GANs~\cite{brock2018large, casanova2021instance, karras2019style} as $G$, due to their sampling speed, and inability to encode real images by design. 
Note that we are not concerned with bad samples, i.e. images with artifacts, as our main goal is to train the image backbone $f$ to map images into features, regardless of image quality. This process is visualized in Fig.~\ref{fig:overview} (b). Also see (a) for a side-by-side comparison of a representative discriminative pretraining paradigm. 

\textit{Encoded dataset} is created by encoding a real image $x$ into the latent space of the generative model using an encoding process to get a latent variable $\tilde{z}$. Then, we similarly run the generative process and record hierarchical intermediate features from $G(\tilde{z})$ to obtain our dataset $D$. This process is visualized in Fig.~\ref{fig:overview}(c). 
Also see Fig.~\ref{fig:ddpm-steps-feat} for encoded ImageNet images and their feature activation maps. 
For generative models that come with an encoder network by design, such as VAEs~\cite{kingma2013auto, vahdat2020nvae}, we can simply re-use it. For diffusion based generative models (DM)~\cite{song2020score, ho2020denoising, dhariwal2021diffusion}, which is the class of generative models we focus our investigation on, we use the forward diffusion process to encode a real image. Specifically, we run forward diffusion for $T$ steps, followed by a single denoising step to extract hierarchical features $f_{l}^{g}$ from intermediate layers of the denoising network, typically a U-Net~\cite{ronneberger2015u}.  See Fig.~\ref{fig:ddpm-steps-feat} for visualization of feature activation maps at different diffusion steps. The choice of $T$ and the encoding process in diffusion models (stochastic~\cite{ho2020denoising} or deterministic~\cite{song2020denoising}) can strongly affect properties of the trained model $f$. We systematically ablate these choices through experiments, and find that distilling stochastically encoded features, which we view as data augmentation in feature space, increases robustness of the downstream backbone $f$.


Both \emph{synthesized} and \emph{encoded} feature datasets can either be pre-computed \textit{offline}, or created \textit{online} while training $f$.
In practice, we use \textit{online sampling} for synthesized datasets, and \textit{online encoding} for encoded datasets to allow fast in-memory access and efficient materialization and removal of samples and corresponding high dimensional features. This allows us to scale to pre-training with datasets and features $\mathbf{f}^g$ of any size without additional pre-processing and storage costs. Online encoding is also the natural choice when using stochastic encoding techniques in diffusion models, since an offline dataset could only store one or a few samples from all possible stochastic encodings of a real image.



\nbf{Feature Regressor} In order to distill generative representations $\mathbf{f}^g$ into a general backbone $f$, we design a feature regressor module that maps and aligns the image backbone's features with the generative features. Inspired by the design of the Feature Pyramid Network (FPN) \cite{lin2017feature}, our feature regressor takes multi-level features from the backbone $f$ and uses a top-down architecture with lateral skip connections to fuse the backbone features and outputs multi-scale features. We apply a Pyramid Pooling Module (PPM) from PSPNet \cite{zhao2017pyramid} similar to \cite{xiao2018unified}, on the last layer of the image backbones before the FPN branch to enhance feature mixing. Fig.~\ref{fig:framework} (bottom) visually depicts this architecture. 

\nbf{Feature Distillation} Denote intermediate features from encoder $f$ at different levels as $\{f_2^e, f_3^e, f_4^e, f_5^e\}$, and the corresponding feature regressor outputs as $\{f_2^r, f_3^r, f_4^r, f_5^r\}$. We use a $1\times1$ convolution to match the number of channels in $f_l^r$ and $f_l^g$, if they are different. Our feature regression loss is simple and is inspired by FitNet~\cite{romero2014fitnets}, which proposed distilling knowledge from a teacher onto a student network by mimicking intermediate feature activations:\\[-3mm]
\begin{equation}\label{eq1}
    \mathcal{L}_{MSE} = \frac{1}{L} \sum_{l}^{L}\Vert f_l^r - \mathbb{W}(f_l^g)\Vert^2_2
\end{equation}
Here, $\mathbb{W}$ is a non-learnable whitening operator implemented as LayerNorm~\cite{ba2016layer}, which normalizes differing feature magnitudes across layers. Layer number $l=\{2,3,4,5\}$ corresponds to features at $2^l$ stride relative to the input resolution.

Additionally, we explore the activation-based Attention Transfer (AT)~\cite{zagoruyko2016paying} objective. AT distills a one dimensional ``attention map" per spatial feature, using an operator defined as  $F_{sum}^p(A)=\sum_{i}^C|A_i|^p$ to sum the power $p$ of the absolute values of the feature activation $A$ across channel dimension $C$,
which improves convergence speed over regressing high dimensional features directly. Specifically,
\begin{equation}
    \mathcal{L}_{AT} = \frac{1}{L} \sum_{l}^{L} \sum_{j \in I}\Vert \frac{Q^r_{l,j}}{\Vert Q^r_{l,j} \Vert _2} - \frac{Q^g_{l,j}}{\Vert Q^g_{l,j} \Vert _2} \Vert _p
\end{equation}
where $Q^r_{l,j} = vec(F_{sum}^p(f^r_{l,j}))$, $Q^g_{l,j} = vec(F_{sum}^p(f^g_{l,j}))$ are respectively the j-th pair in layer $l$ of the regressor's and
generative model's features in vectorized form. We follow \cite{zagoruyko2016paying} to use $p=2$ in our experiments.

Our combined feature regression loss is:
\begin{equation}
    \mathcal{L}_{feat} = \mathcal{L}_{MSE} + \lambda_{AT} \mathcal{L}_{AT}
\end{equation}
where $\lambda_{AT}$ controls the weighting of the loss $\mathcal{L}_{AT}$. We choose $\lambda_{AT}=10.0$ in our experiments, to make the two losses in the same scale. We empirically ablate choices of the loss function and feature regressor designs. 

\subsection{Label-Guided Representation Learning}
\label{sec:labelreps}
In the semi-supervised setting, where a fraction of downstream task labels are available for pre-training, we train a task-dependent branch, called a \emph{feature interpreter}, on top of the frozen generative network $G$ in a supervised manner, following DatasetGAN~\cite{zhang2021datasetgan}. While DatasetGAN synthesized a labeled dataset for training downstream task networks, we instead use soft label distillation for both \emph{encoded} and \emph{synthesized} datasets, i.e. we include predicted soft labels in our feature dataset $D$. This is visualized in Fig.\ref{fig:overview}(d). We first describe the architecture of the \emph{feature interpreter} followed by our distillation objective for soft labels.

\nbf{Feature Interpreter}  We utilize a similar design to BigDatasetGAN~\cite{li2022bigdatasetgan}, which improves the interpreter design over DatasetGAN with better memory efficiency and prediction accuracy. Specifically, the interpreter takes multi-level features $f_l^g$ from the generator as inputs which are fed into a series of \textit{Feature Fusion Layers} (see Fig \ref{fig:framework}) to lower the feature dimension and fuse with the next-level features, to finally output per-pixel logits. We follow BigDatasetGAN's interpreter design and only replace the convolutional fused block with depth-wise separable convolutions~\cite{chollet2017xception}, Group Norm~\cite{wu2018group}, and Swish activation~\cite{ramachandran2017searching}.

We explore training the interpreter branch with segmentation labels, and use a combination of the cross-entropy and Dice~\cite{sudre2017generalised} objectives for training:
\begin{equation}
    \mathcal{L}_{interpreter} = \mathcal{H}(I_{\theta}(f_l^g), y) + \lambda_{d}\mathcal{D}(I_{\theta}(f_l^g), y),
\end{equation}
where  $I_{\theta}$ are the weights of the feature interpreter, $y$ are the task labels. $\mathcal{H}(\cdot,\cdot)$ denotes pixel-wise cross-entropy loss, and $\mathcal{D}(\cdot,\cdot)$ is Dice Loss. $\lambda_d$ is a hyperparameter to weigh the dice loss. We use $\lambda_d=3.0$ in all our experiments following~\cite{sudre2017generalised}.

\nbf{Label Distillation} We follow~\cite{hinton2015distilling} for label distillation. Specifically, we use:
\begin{equation}
    \mathcal{L}_{ld} = \mathcal{H}(P^g_{\tau},P^r_{\tau}), 
\end{equation}
where $P^g_{\tau}$ and $P^r_{\tau}$ are the logits from the feature interpreter 
and the logit-head of the target image backbone $f$, respectively. $\mathcal{H}$ is the cross-entropy, with temperate $\tau$, controlling the sharpness of the output distribution. We use $\tau = 4.0$ in all our experiments following \cite{hinton2015distilling}. 


We use the label distillation objective in conjunction with our feature distillation objective:
\begin{equation}
\mathcal{L}_{mix} = \mathcal{L}_{feat} + \lambda_{ld} \mathcal{L}_{ld}    
\label{eq:mix_distill}
\end{equation}
where $\lambda_{ld}$ is a hyperparameter controlling the weighting between the losses, which we use $\lambda_{ld}=1.0$ in our experiment. We pre-train the image backbone $f$ using the mixed distillation losses over all images in our pre-training dataset, either labeled or unlabeled. Annotated labels are only used for training the feature interpreter, and we only use soft labels from the feature interpreter for pre-training $f$ with distillation.

\begin{figure}[t!]
\vspace{-2mm}
\begin{center}
\includegraphics[width=1.0\linewidth]{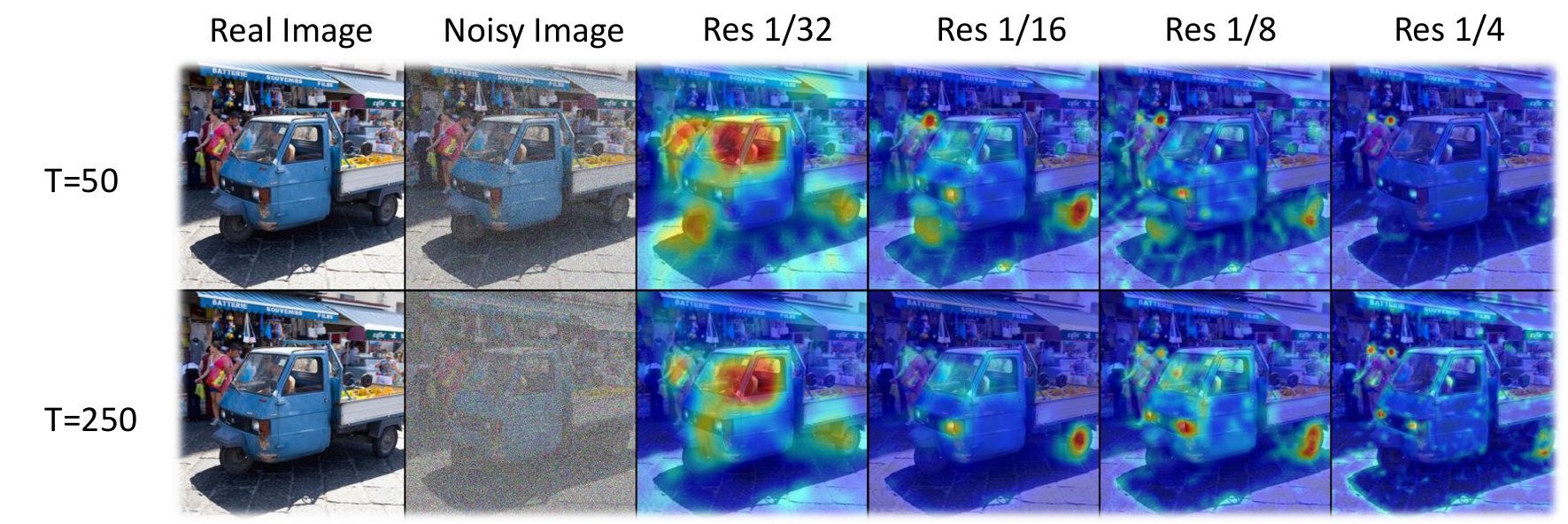}
\end{center}
\vspace{-6mm}
\caption{\textbf{\footnotesize ADM feature visualization (ImageNet)}.
We visualize ADM feature activation maps at different resolution blocks (columns) at different diffusion time steps $T$ 
 (rows). At lower resolution blocks, features activate on objects like humans and cars. For higher resolution block, features focus on smaller parts like wheels and headlights. With increasing $T$, feature activations become smoother. 
}
\label{fig:ddpm-steps-feat}
\vspace{-4mm}
\end{figure}

\begin{table*}[h]
\vspace{-2mm}
\centering
\resizebox{0.8\linewidth }{!}{ 
\setlength{\tabcolsep}{8pt}
\renewcommand{\arraystretch}{1.0}
\begin{tabular}{l|ccc|ccccc}
\toprule
\multirow{2}{*}{Pre-training Method} & PT & \multirow{2}{*}{Arch.}  & Eff. & Cls & \multicolumn{2}{c}{Det.} & \multicolumn{2}{c}{Seg.} \\
& task & & epoch & Acc. & AP$^{bb}$ & AP$^{bb}_{75}$ & AP$^{mk}$ & AP$^{mk}_{75}$\\
\midrule
\rowcolor{LightGray}
\multicolumn{9}{l}{\textit{Vision Transformer Backbone}}\\
Supervised\cite{he2022masked}  & - & ViT-B & 300 & 82.3 & 49.8 & 53.8 & 43.2 & 46.5 \\
MoCov3\cite{chen2021mocov3} & CL & ViT-B & 1600 & 83.2 & - & - & - & - \\
DINO\cite{caron2021emerging} & CL & ViT-B & 1600 & 82.8 & 50.1 & 54.3 & 43.4 & 47.0 \\
BEiT\cite{bao2021beit} & MIM & ViT-B & 800 & 83.2 & 50.1 & 54.6 & 43.5 & 47.1 \\
MAE\cite{he2022masked} & MIM & ViT-B & 1600 & 83.6 & - & - & - & - \\
iBOT\cite{zhou2021ibot}  & MIM + CL & ViT-B  & 1600 & 84.0 & 51.2 & 55.5 & 44.2 & 47.7 \\
\midrule
\rowcolor{LightGray}
\multicolumn{9}{l}{\textit{Convolutional Backbone}}\\
Supervised\cite{liu2022convnet} & - & ConvX-B & 300 & 83.8 & 51.2 & 55.5 & 44.3 & 47.9 \\
SparK\cite{tian2023designing} & MIM & ConvX-B & 1600 & \textbf{84.8} & 51.9 & 56.5 & 44.6 & 48.4 \\
\rowcolor{LightBlue}
DT-feat.distil. w/ ADM\cite{dhariwal2021diffusion} & GEN & ConvX-B & *600 & 83.9 & \textbf{52.5} & \textbf{57.4} & \textbf{45.2} & \textbf{49.0}\\
\bottomrule
\end{tabular}}
\vspace{-2mm}
\caption{
\textbf{\footnotesize Comparing DreamTeacher with SoTA self-supervised methods on ImageNet and instance segmentation on COCO.} All the baselines including ADM are pre-trained on ImageNet-1k. For ImageNet classification, we adopt SparK's fine-tuning setting with resolution 224. For COCO, we follow iBOT to fine-tune Cascade Mask R-CNN\cite{cai2019cascade} for 12 (1$\times$) epochs. Average precisions of detection box (AP$^{bb}$) and segmentation mask (AP$^{mk}$) on val2017 are reported. 
For a fair comparison, both our method and baselines follow iBOT fine-tuning schedule and setting. Our DT pre-training task is highlighted as generative(GEN) comparing to contrastive(CL) and masking(MIM) based objectives. *Our effective epochs includes 400 epochs generative model training and 200 epochs feature distillation training.}

\vspace{-1mm}
\label{convNX}
\end{table*}
\begin{table*}[h]
\vspace{-0.5mm}
\centering
\begin{minipage}{0.55\textwidth}
\resizebox{1.0 \textwidth }{!}{ 
\setlength{\tabcolsep}{4pt}
\renewcommand{\arraystretch}{1.0}
\begin{tabular}{l|cc|ccccc}
\toprule
\multirow{2}{*}{Pre-training (ResNet-50)}& PT & Eff. & Cls. & \multicolumn{2}{c}{1$\times$ 
 Schedule} & \multicolumn{2}{c}{2$\times$ Schedule}\\
 & task & epoch & (Acc.) & AP$^{bb}$ &  AP$^{mk}$ & AP$^{bb}$ &  AP$^{mk}$\\
\midrule
Supervised & - & - & 79.8 & 38.9 & 35.4 & 41.3 & 37.3\\
SimSiam\cite{chen2021exploring} & CL & 800 & 79.1 & - & - & - & -\\
MoCo\cite{he2020momentum} & CL & 800 & - & 38.5 & 35.1 & 40.8 & 36.9\\
MoCov2\cite{chen2020improved} & CL & 1600 & 79.8 & 40.4 & 36.4 & 41.7 & 37.6\\
SimCLR\cite{chen2020simple} & CL & 4000 & 80.0 & - & - & - & -\\
InfoMin\cite{tian2020makes} & CL & 800 & - & 40.6 & 36.7 & 42.5 & 38.4\\
BYOL\cite{grill2020bootstrap} & CL & 1600 & 80.0 & 40.4 & 37.2 & 42.3 & 38.3\\
SwAV\cite{caron2020unsupervised} & CL & 1200 & 80.1 & - & - & 42.3 & 38.2\\
SparK\cite{tov2021designing} & MIM & 1600 & \textbf{80.6} & 41.6 & 37.7 & 43.4 & 39.4\\
\rowcolor{LightBlue}
DT-feat.distil. w/ ADM\cite{dhariwal2021diffusion} & GEN & *600 & 80.2 & \textbf{44.1} & \textbf{40.1} & \textbf{45.1} & \textbf{40.8}\\
\bottomrule
\end{tabular}}
\vspace{-2mm}
\caption{
\textbf{\footnotesize ResNet-50 results on ImageNet and COCO instance segmentation.} For ImageNet classification, we follow SparK's fine-tuning setting with resolution 224. Top-1 accuracy (Acc) on ImageNet val set is reported. For COCO, Mask R-CNN\cite{he2017mask} ResNet50-FPN is equally fine-tuned for 12 or 24 epochs (1$\times$ or 2$\times$), following the same setup as SparK.  *Our effective epochs includes 400 epochs generative model training and 200 epochs feature distillation training.
}
\label{resnet-coco}
\end{minipage}
\hfill
\begin{minipage}{0.4\textwidth}
\resizebox{1.0\linewidth }{!}{ 
\setlength{\tabcolsep}{4pt}
\renewcommand{\arraystretch}{1.0}
\begin{tabular}{l|ccc}
\toprule
\multirow{2}{*}{Pre-training (ResNet-50)}&  \multicolumn{1}{c}{ADE20k} & \multicolumn{2}{c}{BDD100k}\\
& mIoU & AP$^{bb}$ & AP$^{mk}$\\
\midrule
Supervised  & 40.9 & 26.1 & 20.2\\
SimCLR\cite{chen2020simple} & 39.9 & 24.5& 20.6\\
SparK\cite{tov2021designing} & 40.5 & 25.7 & 22.4\\
SimSiam\cite{chen2021exploring} & 40.6 & 26.3& 22.7\\
MoCov2\cite{chen2020improved} & 40.9 & 26.9& 22.9\\
denseCL\cite{wang2021dense} & 41.1 & 27.1& 23.4 \\
SwAV\cite{caron2020unsupervised}& 41.2 & 25.6&22.2\\
BYOL\cite{grill2020bootstrap} & 41.6 & 26.2&22.6\\
PixPro\cite{xie2021propagate} & 41.6 & 27.2&23.1\\

\rowcolor{LightBlue}
DT-feat.distil. w/ ADM\cite{dhariwal2021diffusion} & \textbf{42.5}& \textbf{28.3} & \textbf{24.8}\\

\bottomrule
\end{tabular}}
\vspace{-2mm}
\caption{
\textbf{\footnotesize Transfer learning: ADE20k and BDD100k.} All methods are pre-trained on ImageNet-1k and fine-tuned on downstream tasks. For ADE20k, we follow \cite{liu2022convnet} to use UperNet\cite{xiao2018unified} and fine-tune for 160k iterations, reported number is mean IoU at single scale. For BDD100k, we follow official setup\cite{yu2020bdd100k} to use Mask R-CNN ResNet50-FPN fine-tune for 36 (3$\times$) epochs.
} 
\label{resnet-ade}
\end{minipage}
\vspace{-4mm}
\end{table*}
\vspace{-2mm}
\section{Experiments}
\label{sec:experiment}
\vspace{-2mm}
In this section, we first experimentally evaluate the performance of {\ourmodel} for both: self-supervised representation learning and semi-supervised learning (Subsec.~\ref{sec:imagenet-transfer}). 
We then additionally investigate the performance of our model for \textit{in-domain-pretraining} (Subsec.~\ref{sec:indomain}). In the \textit{in-domain} setting, the same target dataset is used for both pretraining and finetuning, and the backbones are initialized from scratch. Finally, we ablate different generative models and design choices of DreamTeacher (Subsec.~\ref{sec:ablation}).

We investigate several generative models: for GANs, we use unconditional BigGAN~\cite{brock2018large}, ICGAN~\cite{casanova2021instance}, StyleGAN2~\cite{karras2020analyzing} and for diffusion-based model, ADM~\cite{dhariwal2021diffusion}, and Stable Diffusion (SD) Models\cite{rombach2021highresolution}.
We use four datasets for pre-training, both for training the generative models, as well as knowledge distillation to downstream backbones. We use \textit{BDD100K}~\cite{yu2020bdd100k}, 
\textit{ImageNet-1k(IN1k-1M)}, \textit{LSUN}~\cite{yu2015lsun} and  \textit{FFHQ}~\cite{karras2019style}, which contain 100k, 1.28 million, 10 million, and 100k images, respectively. We focus on convolutional networks as target image backbones.

\subsection{ImageNet Evaluation and Transfer}
\label{sec:imagenet-transfer}
\nbf{Imagenet Pretraining}
We first validate the effectiveness of {\ourmodel}  for ImageNet pretraining. In this setting, we follow the recent SoTA method, SparK~\cite{tian2023designing}, and evaluate two convolutional architectures as downstream backbones, ConvNext-B\cite{liu2022convnet} and ResNet50\cite{he2016deep}. Following common practice in the literature~\cite{he2022masked, tian2023designing}, we pre-train image backbones unsupervised on ImageNet-1k.  
For a comparison with transformer-based self-supervised methods, we follow SparK's methodology\cite{tian2023designing} and pre-train a modern CNN-based backbone ConvNeXt\cite{liu2022convnet} with a similar number of parameters. Additionally, to ensure a fair comparison with CNN-based self-supervised methods, we pre-train and evaluate a classical backbone, ResNet-50. 

\nbf{Implementation} We use pre-trained unconditional ADM with resolution 256 from the official release. We only use horizontal flip augmentation and train using LAMB \cite{you2019large} optimizer with a batch size of 2048. We adopt a cosine-annealing learning rate with peak value = 0.0002 $\times$ \textit{batchsize} / 256. See appendix for other hyperparameters.

\nbf{Transferring to Downstream Tasks} We assess the quality of learned representations obtained using {\ourmodel} by fine-tuning the pre-trained backbone with additional heads per task (see Appendix for implementations). We test downstream transfer performance for ImageNet classification and COCO~\cite{lin2014microsoft} instance segmentation, which are representative global and spatial image understanding tasks commonly used in literature.
Prior self-supervised learning methods have excelled at ImageNet classification, and have recently shown improvement over supervised ImageNet pre-training for spatial understanding tasks such as object detection and segmentation that are much more cost-intensive to label. Additionally, we also include linear probing experiments on ImageNet for both classification and semantic segmentation tasks in the Appendix (Table~\ref{in_linear}).


\nbf{Discussion} Comparing to self-supervised methods based on vision-transformer, {\ourmodel} outperforms existing approaches in both detection and segmentation, and performs on par in the classification setting (Table~\ref{convNX}). Specifically,
 {\ourmodel} achieves 52.5 AP$^{bb}$ and 45.2 AP$^{mk}$ on the COCO instance segmentation task 
 outperforming the SoTA transformer-based method iBOT by $+1.3$ and $+1.0$. 
%
{\ourmodel} also outperforms the recently proposed sparse-convolution based MIM method SparK \cite{tian2023designing}, in the tasks of detection and segmentation by $+0.6$ and $+0.6$, respectively. We notice that our method does not outperform this baseline on the task of image classification. This may likely be due to our approach of distilling spatial features from the generative model, which might contain more semantically localized information for generation (visualized in Fig.~\ref{fig:ddpm-steps-feat}), which empirically seems to favor dense prediction tasks.
It is also worth noting that our method is $\sim 2.5 \times$ more efficient than SparK w.r.t. effective training epochs~\cite{tian2023designing} on ImageNet (600 vs 1600). This number includes training steps of the generative model, ADM.

 In Table~\ref{resnet-coco} we show results for Resnet-50 using SparK's setting and parameters. Specifically, we evaluate ImageNet classification performance with full fine-tuning and COCO instance segmentation with two schedules (1$\times$ and 2$\times$). 
 Similar to the previous experiment, we achieve comparable performance as baselines 
 for ImageNet classification. 
 For COCO instance segmentation, we notably outperform all contrastive methods and the masking-based approach SparK ($+2.5$ AP$^{bb}$ for 1$\times$ and $+1.7$ AP$^{bb}$ for 2$\times$ schedule). In Table~\ref{resnet-ade}, we further evaluate transfer learning on the ADE20k semantic segmentation task and BDD100k instance segmentation task. We include SoTA contrastive methods for dense prediction tasks, denseCL\cite{wang2021dense} and PixPro\cite{xie2021propagate}. Our approach using generation as pre-training task outperforms both global and dense contrastive pre-training tasks as well as the masked image modelling task.

\begin{table*}[t!]
\vspace{-2mm}
\begin{minipage}{0.36\textwidth}
\centering
\resizebox{1.0\linewidth }{!}{ 
\setlength{\tabcolsep}{4pt}
\renewcommand{\arraystretch}{1.0}
\begin{tabular}{l|cc|cc}
\toprule
\multirow{2}{*}{Pre-training (ResNet-50)} & PT & Eff.  & \multicolumn{2}{c}{BDD100k Ins.}\\
& task & epoch & AP$^{bb}$ & AP$^{mk}$ \\
\midrule
Supervised\cite{yu2020bdd100k} & - & - & 26.1 & 20.2 \\
BYOL\cite{grill2020bootstrap} & CL & 5000 & 23.9 & 20.0\\
SparK\cite{tian2023designing} & MIM & 2500 & 24.4 & 20.6\\
\rowcolor{LightBlue}
DT-feat.distil. w/ StyleGAN2\cite{karras2020analyzing} & GEN & *900  & 25.1 & 21.4\\
\rowcolor{LightBlue}
DT-feat.distil. w/ ADM\cite{dhariwal2021diffusion} & GEN & *900  & \textbf{26.7} & \textbf{22.9}\\

\bottomrule
\end{tabular}}
\vspace{-2mm}
\caption{
\textbf{\footnotesize In-domain pre-training on BDD100k.} We follow the recommendation of \cite{el2021large} to pre-train contrastive and masking based self-supervised method with long schedule for small dataset like BDD100k with 70k train images. We finetune on BDD100k instance segmentation task using Mask R-CNN ResNet50-FPN for 36(3$\times$) epochs. 
}
\label{bdd_indomain}
\end{minipage}
\hfill
\begin{minipage}{0.60\textwidth}
\centering
\resizebox{1.0\textwidth }{!}{ 
\setlength{\tabcolsep}{2pt}
\renewcommand{\arraystretch}{1.0}
\begin{tabular}{lccccccc}
\toprule
method & backbone & params & pre-data & Bedroom-28 & FFHQ-34 & Cat-15 & Horse-21\\
\midrule
classific. sup. & RN101 & 43M & IN1k-1M & 34.4 & 53.6 & 38.8 & 51.1 \\
classific. sup. & ConvNX-B & 89M & IN21k-14M & 41.0 & 59.2 & 47.3 & 56.0 \\
\midrule
SwAV\cite{caron2020unsupervised} & RN50-w2 & 94M& task domains & 41.0 & 54.7 & 44.1 & 51.7  \\
MAE\cite{he2022masked} & ViT-L & 305M & task domains & 45.0 & 58.8 & 52.4 & 63.4  \\
DatasetGAN\cite{zhang2021datasetgan} & RN101 & 43M & task domains & 31.3 & 57.0 & 36.5 & 45.4  \\
DatasetDDPM\cite{baranchuk2021labelefficient}  & RN101 & 43M & task domains & 47.9 & 56.0 & 47.6 & 60.8 \\
DDPM-seg\cite{baranchuk2021labelefficient}   & UNet & 554M & task domains & 49.4  & 59.1 & 53.7 & 65.0 \\

\rowcolor{LightBlue}
DT-mix.distil. w/ ADM\cite{dhariwal2021diffusion}& RN101 & 43M & task domains&  49.9  &  59.4 & 56.7 &  65.9 \\
\rowcolor{LightBlue}
DT-mix.distil. w/ ADM\cite{dhariwal2021diffusion}& ConvNX-B & 89M & task domains& \textbf{54.8}  & \textbf{61.2} & \textbf{58.6} & \textbf{67.6} \\
\bottomrule
\end{tabular}}
\vspace{-2.5mm}
\caption{
\textbf{Label-efficient semantic segmentation benchmark.} We compare our {\ourmodel} (DT) with various representation learning baselines. 
Our \emph{DT-mix.distil.} with ResNet 101 backbone (only 43M parameters) beats all baselines, some with 10x the number of parameters. We also show our method with ConvNX-B achieves the new SoTA without using any extra data, i.e. IN1k-1M or IN21k-14M.  
} 
\label{label-efficient}
\end{minipage}
\vspace{-4mm}

\end{table*}

\vspace{-1mm}
\subsection{In-domain Pre-training}
\label{sec:indomain}
\vspace{-1mm}
For in-domain pre-training, we first pre-train the backbone with various self-supervised training approaches. 
Pre-training efficacy is evaluated by fine-tuning the backbone on different tasks with label supervision, on the same dataset. 
Note that both baselines and {\ourmodel} use randomly initialized downstream backbones. We evaluate unsupervised pre-training using the BDD-100k benchmark and semi-supervised pre-training using multiple datasets from the label efficiency benchmark used by~\cite{zhang2021datasetgan, baranchuk2021labelefficient, li2022bigdatasetgan}. 

\nbf{BDD100k Benchmark} We pre-train all self-supervised learning methods, including {\ourmodel} on 70k unlabeled images from BDD100k. We then evaluate all methods on BDD100k, which contains 10k images annotated with semantic, instance and panoptic labels. We follow the official dataset split, using 7k labels for supervised training. Results are reported on the validation set (1k images). We use a Resnet-50~\cite{he2016deep} backbone for all methods.
\begin{figure}[t!]
\vspace{-2mm}
\begin{center}
\includegraphics[width=0.95\linewidth]{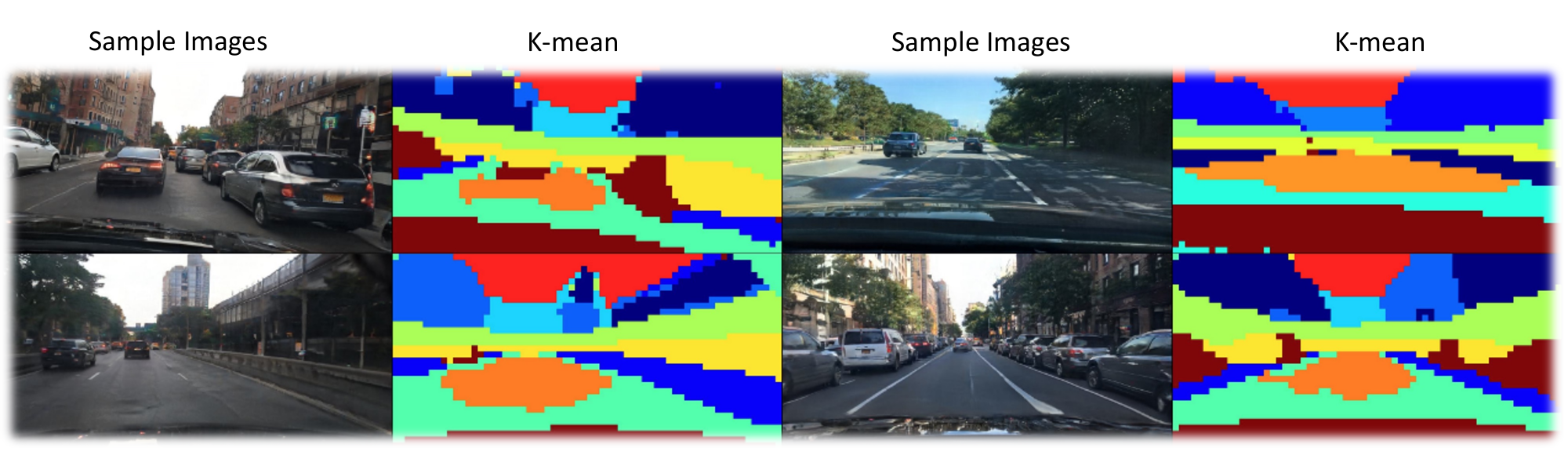}
\end{center}
\vspace{-6mm}
\caption{\textbf{\footnotesize K-means clustering of StyleGAN2's features trained on BDD100K.} We run kmeans clustering ($k=10$) on 10k sampled features, and show unsupervised segmentation maps on sampled images. Notice that the clusters are consistent across images (car, sky, tree etc), indicating a semantic meaning of the generative features.
}
\label{fig:stylegan-kmean}
\vspace{-6mm}
\end{figure}

\nbf{Feature Visualization} We visualize the knowledge learned by different generative models in Fig.~\ref{fig:stylegan-kmean}. Specifically, we show scenes sampled from StyleGAN2 trained on BDD100k. We perform k-means clustering ($k=10$) of StyleGAN's features and visualize clusters with different colors. Notice that the clusters roughly correspond to major semantic classes. 

\nbf{Results} In Table~\ref{bdd_indomain}, we compare {\ourmodel} with the representative contrastive method BYOL and recently proposed MIM-based method SparK on BDD100k instance segmentation task. As investigated in\cite{el2021large}, contrastive and masking-based self-supervised methods require a longer pre-training schedule to converge on a small in-domain dataset. 
We pre-trained backbones using DreamTeacher feature distillation with StylegGAN2 and ADM, and the effective epochs comprise 300 training epochs of the generative model and 600 training epochs for feature distillation.
Our methods outperform contrastive and masking-based techniques significantly for in-domain pre-training with better training efficiency. Notably, our method with ADM outperforms the ImageNet supervised pre-trained backbone, showing promising results without relying on large-scale curated datasets like ImageNet. 
See appendix for qualitative results and semantic segmentation and panoptic segmentation results.

\nbf{Label-efficient Benchmarks} We now evaluate in-domain pre-training in our semi-supervised setting. We follow the setup in DDPM-seg~\cite{baranchuk2021labelefficient} and train on ``bedroom", ``cat" and ``horse" categories
from LSUN~\cite{yu2015lsun}, and human faces from FFHQ ~\cite{karras2019style} (at 256x256 resolution). We evaluate semantic segmentation, where the datasets have 28, 15, 21, 34 semantic classes, respectively. Datasets contain only 40, 30, 30 and 20 labeled images. 
We pre-train all backbones from scratch, i.e. without ImageNet pre-trained initialization.
We use UPerNet~\cite{xiao2018unified} for semantic segmentation. Note that some baselines utilize different settings. DatasetGAN~\cite{zhang2021datasetgan} and DatasetDDPM~\cite{baranchuk2021labelefficient} both train a small task-specific head on top of a pre-trained generative model, and generate a large labeled dataset for training a downstream network. On the other hand, DDPM-seg directly leverages the diffusion-based generative model with a task head as the segmentation network.\looseness=-1

Results are reported in Table~\ref{label-efficient}
. We highlight several key observations below:\\[-6mm]
\begin{itemize}[leftmargin=*]
 \item Given the same backbone, ResNet101, {\ourmodel} trained with our mixed distillation (Eq.~\ref{eq:mix_distill}) outperforms DatasetDDPM across all datasets. We outperform DatasetDDPM by 3.4\% on FFHQ-34, and 9.1\% on Cat-15. \\[-6mm]
  \item  Using both a 10x and a 6x smaller backbone (ResNet-101 and a ConvNX-B~\cite{liu2022convnet}, respectively), we outperform DDPM-Seg on all classes. On Bedroom-28 and Cat-15, we  improve over the baseline by more than 5\%
  .\\[-6mm]
  \item  Given the same backbone, our method significantly outperforms pre-training with ImageNet classification labels. With ConvNX-B~\cite{liu2022convnet}, our proposed approach is better than ImageNet pre-training by more than 10\% on Bedroom-28, Cat-15, and Horse-21. 
  These results may indicate that if the in-domain datasets are sufficiently large relative to the complexity of the task, in-domain pre-training is more effective than pre-training on large generic datasets like ImageNet. 
  Note that this is true for both semi-supervised (these results) and unsupervised pre-training (Table~\ref{bdd_indomain}).
\end{itemize}

\vspace{-2mm}
\subsection{Ablation Studies}
\label{sec:ablation}
We first ablate DreamTeacher with different generative models in Table~\ref{ablate_gen}.
Result shows ADM trained on IN1k-1M has the highest downstream performance. We also exploited off-the-shelf Stable Diffusion trained on LAION-400M and pretrain backbone on IN1k-1M. However, it performs slightly worse than ADM trained on IN1k-1M. 
In Table~\ref{abl:distillation_loss} we ablate our proposed distillation losses.  Mixing feature- and label- distillation achieves the best performance except for the FFHQ-34 dataset. 
We demonstrate our design choices of the decoder used in pre-training in Table~\ref{tab:ablate-regressor}, loss functions in Table~\ref{tab:ablate-loss}, encoding modes (deterministic and stochastic) in Table~\ref{tab:encoding_mode} and diffusion steps in Table~\ref{tab:noise_steps}. Ablation studies pre-train backbone for 100 epochs and report performance on BDD100k instance segmentation. These results confirm our choices.

\begin{table}[t!]
\vspace{-2mm}
\centering
\resizebox{1.0\linewidth }{!}{ 
\setlength{\tabcolsep}{4pt}
\renewcommand{\arraystretch}{1.0}
\begin{tabular}{lcc|ccc}
\toprule
\multirow{2}{*}{Pre-training (ResNet-50)} & \multirow{2}{*}{Gen. Data} &  \multirow{2}{*}{Pre. Data} & \multicolumn{1}{c}{ADE20k} & \multicolumn{2}{c}{COCO}\\
& & & mIoU & AP$^{bb}$ & AP$^{mk}$ \\
\midrule

DT-feat.distil. w/ BigGAN\cite{brock2018large}   & IN1k-1M & IN1k-1M &40.8 & 40.7 & 36.9\\
DT-feat.distil. w/ ICGAN\cite{casanova2021instance}  & IN1k-1M &  IN1k-1M & 41.2 & 40.0 & 36.5\\
DT-feat.distil. w/ SD1.4\cite{rombach2021highresolution} & LAION-400M &  IN1k-1M & 41.4 & 43.3 & 39.4\\
DT-feat.distil. w/ ADM\cite{dhariwal2021diffusion} & IN1k-1M &   IN1k-1M & 42.5 & 44.1 & 40.1\\
\bottomrule
\end{tabular}}
\vspace{-3mm}
\caption{
\textbf{\footnotesize Ablation study with different generative models using DreamTeacher.} We use off-the-shelf SD with version 1.4 pre-trained on LAION-400M without finetuning, and it performs slightly worse than DT with ADM, which is trained on ImageNet-1k.
} 
\vspace{-1mm}
\label{ablate_gen}
\end{table}

\begin{table}[t!]
\vspace{-2mm}
\centering
\resizebox{1.0\linewidth }{!}{ 
\setlength{\tabcolsep}{8pt}
\renewcommand{\arraystretch}{1.0}
\begin{tabular}{lcccc}
\toprule
Loss  & Bedroom-28 & FFHQ-34 & Cat-15 & Horse-21\\
\midrule
feat distil. & 53.1 & 61.1 & 58.2 & 64.7 \\
label distil. & 54.6  & 61.3 & 58.4 & 64.4 \\
mix distil. & 54.8  & 61.2 & 58.6 & 67.6 \\
\bottomrule
\end{tabular}}
\vspace{-2mm}
\caption{
\textbf{\footnotesize Ablating feature/label distillation.} We pretrain ConvNeXt-B to convergence. Feature distillation (FD) does not leverage labels in pre-training, yet performs competitively. 
%
} 
\label{abl:distillation_loss}
\vspace{-1mm}
\end{table}

\begin{table}[t!]
\vspace{-2mm}
\centering
\begin{minipage}{0.22\textwidth}
    \resizebox{1\textwidth }{!}{ 
    \setlength{\tabcolsep}{2pt}
    \renewcommand{\arraystretch}{1.0}
    \begin{tabular}{lcc}
    \toprule
    Decoder & \textit{Box mAP} & \textit{Mask mAP} \\
    \midrule
    FPN & 23.6 & 20.3   \\
    FPN+Atten. layer & 23.9 & 20.7 \\
    PaFPN & 24.0 & 20.8   \\
    FPN+PPM & 25.1 & 21.4   \\
    \bottomrule
    \end{tabular}}
    \vspace{-2mm}
    \caption{
    \textbf{\footnotesize Ablating feat. regressors} We pretrain ResNet50 with 
    FT. We compare FPN with an attention layer, and add a bottom-up branch to fuse FPN features (PaFPN). 
    } 
    \label{tab:ablate-regressor}
\end{minipage}
\hfill
\begin{minipage}{0.22\textwidth}
    \vspace{-4mm}
    \centering
    \resizebox{1\linewidth}{!}{ 
    \setlength{\tabcolsep}{1pt}
    \renewcommand{\arraystretch}{1.0}
    \begin{tabular}{lcc}
    \toprule
    Decoder & \textit{Box mAP} & \textit{Mask mAP} \\
    \midrule
    Finnet(MSE) & 23.6 & 20.3 \\
    AT & 22.3 & 19.3 \\
    MSE+AT & 25.0 & 21.6\\
    \bottomrule
    \end{tabular}}
    \vspace{-2mm}
    \caption{
    \textbf{\footnotesize Ablating distillation losses.} We pretrain ResNet50 with MSE or AT loss using feature distill. Combining losses achieves best results.
    } 
    \label{tab:ablate-loss}
\end{minipage}
\hfill
\begin{minipage}{0.22\textwidth}
    \resizebox{1\textwidth }{!}{ 
    \setlength{\tabcolsep}{1pt}
    \renewcommand{\arraystretch}{1.0}
    \begin{tabular}{lcc}
    \toprule
    Encoding & \textit{Box mAP} & \textit{Mask mAP} \\
    \midrule
    Determin. & 23.4 & 20.8   \\
    Stochastic & 24.3 & 21.1 \\
    \bottomrule
    \end{tabular}}
    \vspace{-2mm}
    \caption{
    \textbf{\footnotesize Ablating DDPM encoding.} We use DDIM~\cite{song2020denoising} sampling for deterministic encoding. 
    In both cases, backbone is pretrained for 100 epochs. 
    } 
    \label{tab:encoding_mode}
\end{minipage}
\hfill
\begin{minipage}{0.22\textwidth}
    \vspace{-2mm}
    \centering
    \resizebox{1\linewidth}{!}{ 
    \setlength{\tabcolsep}{3pt}
    \renewcommand{\arraystretch}{1.0}
    \begin{tabular}{lcc}
    \toprule
    Steps & \textit{Box mAP} & \textit{Mask mAP} \\
    \midrule
    T=50 & 23.8 & 20.4   \\
    T=150 & 23.9 & 20.6 \\
    T=250 & 24.4 & 21.1 \\
    T=350 & 23.4 & 20.1 \\
    \bottomrule
    \end{tabular}}
    \vspace{-2mm}
    \caption{
    \textbf{\footnotesize Ablating \# of diffusion steps.} We pretrain ResNet50 with feature distillation using  different \# of diffusion steps. Performance varies with T. 
    } 
    \label{tab:noise_steps}
\end{minipage}
\vspace{-4mm}
\end{table}

\vspace{-5mm}
\paragraph{Limitations:} Our framework relies on generative models for representation learning, and training a generative model on large-scale datasets at  high resolution is costly, especially with diffusion-based models. Further, our feature distillation method only considers features at the same spatial resolution and we limit our scope to CNN-based image backbones. 
Distilling features into vision transformers is for future work. 

\vspace{-6mm}
\section{Conclusion}
\label{sec:conclusion}
\vspace{-1mm}
We proposed {\ourmodel}, a framework for distilling knowledge from generative models onto target image backbones. We investigated several different settings, generative models, target backbones, and benchmarks. Experiments show that generative networks that leverage large unlabeled datasets with generative objectives learn semantically meaningful features that can be successfully distilled on target image backbones. We empirically show our generative-based pre-training method outperforms existing contrastive based and MIM based self-supervised learning approaches in several challenging benchmarks including COCO, ADE20K and BDD100K. We hope our exploration and discovery can inspire future works to study generative pre-training and leveraging geneartive models for vision tasks.


{\small
\bibliographystyle{ieee_fullname}
\bibliography{11_references}
}

\ifarxiv \clearpage In the Appendix we provide additional studies, implementation details, and qualitative results.
\appendix
\label{sec:appendix}
\section{ImageNet Linear Evaluation}
We show the results obtained on ImageNet with self-supervised pre-training, via linear probing as well as ImageNet foreground segmentation readout results in Table~\ref{in_linear}.

\nbf{Linear probing} We follow SimSiam~\cite{chen2021exploring} setup. Specifically,  given the pre-trained network, we train a supervised linear classifier on frozen features from the ResNet's global average pooling layer. When training the linear classifier we use a learning rate of $lr=0.02$ with a cosine decay schedule for 90 epochs, weight decay = 0, momentum = 0.9, a batch size of 4096, and a LARS optimizer \cite{you2017large}. After training the linear classifier, we evaluate it on the centered 224x224 crop in the validation set.

\nbf{Semantic segmentation readout} We also evaluate the pre-trained backbone's features on the dense prediction tasks on ImageNet. We use the dataset from BigDatasetGAN~\cite{li2022bigdatasetgan}, which has object mask annotation on ImageNet 1k classes. On average, each class has 5 annotated images and in total, it has 6,571 training images and 1,813 testing images. We use the dataset to evaluate the pre-trained backbones on foreground/background semantic segmentation performance. We train an FCN~\cite{long2015fully} decoder with the frozen pre-trained backbone's features. The FCN readout network is trained using the SGD optimizer with $lr=0.01$, momentum=0.9 and weight decay=0.0005. We use poly learning rate schedule with power=0.9 and train for 20k iterations. Both training and testing images are center cropped with 256x256 resolution. For all methods, we use ResNet-50 as the image backbone. 

\nbf{Results} Our method with feature distillation and the ADM generative model~\cite{dhariwal2021diffusion}, referred to as \emph{DT-feat.distil w/ ADM} achieves 63.9 Top1 accuracy with linear probing, outperforming all generative pre-trained baselines, and also outperforms competitive discriminative pre-train backbones SimCLR and denseCL. However, we do not outperform the best performing contrastive based method BYOL in this task. Potential explanation could be our pre-training focus on spatial features instead of global discriminative features, which is important for global classification tasks. Notably, our method achieves 79.3 mIoU on the ImageNet foreground/background segmentation task (readout), outperforming all baselines, including dense contrastive based method denseCL, showing the advantage of our method for downstream dense prediction tasks.

\begin{table}[t!]
\centering
\resizebox{0.5\textwidth }{!}{ 
\setlength{\tabcolsep}{4pt}
\renewcommand{\arraystretch}{1.0}
\begin{tabular}{lc|cc}
\toprule
\multirow{2}{*}{Pre-training (ResNet-50)} & Eff. & \multicolumn{1}{c}{\textbf{ImageNet Cls.}} & \multicolumn{1}{c}{\textbf{ImageNet Seg.}}\\
& epoch& \textit{Linear Top1} & \textit{Readout mIoU}\\
\midrule
Supervised  & - & 79.3 & 74.1 \\
\midrule
\rowcolor{LightGray}
\multicolumn{4}{l}{\textit{Discriminative Pretrain:}}\\
SimCLR\cite{chen2020simple}  & 4000&62.5 & 73.9\\
denseCL\cite{wang2021dense} & 1600& 63.6 & 76.5\\
MocoV2\cite{chen2020improved}  & 1600& 67.5 & 76.3\\
SimSiam\cite{chen2021exploring} & 800& 69.8 & 75.0\\
SwAV\cite{caron2020unsupervised}  & 1200& 70.4 & 76.1\\
BYOL\cite{grill2020bootstrap} &  1600& \textbf{71.7} &  75.9\\
\rowcolor{LightGray}
\multicolumn{4}{l}{\textit{Generative Pretrain:}} \\
BiGAN\cite{donahue2016adversarial} & -& 31.0 & -\\
BigBiGAN\cite{donahue2019large} & -& 56.6 & -\\
SparK\cite{tov2021designing} & 1600& 54.1 & 75.6 \\
\rowcolor{LightBlue}
DT-feat.distil w/ ADM\cite{dhariwal2021diffusion}  &*600& 63.9 & \textbf{79.3}\\
\bottomrule
\end{tabular}}

\caption{
\textbf{ImageNet linear classificiation and semantic segmentation performance.} Our method with ADM includes 400 epochs of generative model pre-training and 200 epochs of feature distillation onto the backbone.
} 
\label{in_linear}
\vspace{-3mm}
\end{table}

\section{BDD100K Pre-training}
We show in-domain pre-training on BDD100K instance segmentation task in the main text. Here we show in-domain transfer learning results on two more tasks: semantic segmentation and panoptic segmentation on BDD100K. We further show transfer learning performance on Cityscapes.

\nbf{In-domain transferring} Following the setup in the main text, we investigate in-domain pre-training in the driving dataset BDD100K. We pre-trained backbones of all methods from scratch using 70k unlabeled images in BDD100K training set. Following the recommendation of \cite{el2021large}, we pre-trained contrastive based method BYOL and MIM based method SparK with longer pre-training epochs. We then fine-tune the pre-trained backbone on 7k labeled training dataset on semantic segmentation and panoptic segmentation tasks. In Table~\ref{bdd_indomain_pan}, we show DreamTeacher using StyleGAN2 and ADM outperform BYOL and SparK pre-training on semantic segmentation and panoptic segmentation task, achieving 60.3 mIoU and 21.8 PQ. However, it does not outperform ImageNet supervised pre-training. One reason can be the unlabelled driving dataset is small and lacks of rare objects and concepts comparing to well-curated object-centred dataset ImageNet.

\nbf{Cityscapes transferring} Next, we evaluate pre-training on BDD100K~\cite{yu2020bdd100k}, and transfer to the Cityscapes~\cite{cordts2016cityscapes} semantic segmentation benchmark, which contains 2,975 training images. While both are autonomous driving datasets, they have a domain gap, being captured with different cameras and in different cities/continents. Table~\ref{bdd_transfer} shows that our method outperforms all self-supervised baselines significantly. Compared to the best-performing baseline, SparK, our performance gain increases when the number of available downstream labeled examples decreases. We outperform SparK by 1.8\%, 2.7\%, 4.0\% and 4.7\%, when fine-tuning on 2,975, 744, 374 and 100 labeled examples. 

\begin{table}[t!]
\vspace{-2mm}
\centering
\resizebox{1.0\linewidth }{!}{ 
\setlength{\tabcolsep}{4pt}
\renewcommand{\arraystretch}{1.0}
\begin{tabular}{l|cc|cc}
\toprule
\multirow{2}{*}{Pre-training (ResNet-50)} & PT & Eff.  & Seg. &Pan.\\
& task & epoch & mIoU & PQ \\
\midrule
Supervised\cite{yu2020bdd100k} & - & - & 61.1 & 22.4 \\
\midrule
BYOL\cite{grill2020bootstrap} & CL & 5000 & 58.4 & 20.9\\
SparK\cite{tian2023designing} & MIM & 2500 & 56.9&20.2 \\
\rowcolor{LightBlue}
DT-feat.distil. w/ StyleGAN2\cite{karras2020analyzing} & GEN & *900  & 59.7 & 21.5\\
\rowcolor{LightBlue}
DT-feat.distil. w/ ADM\cite{dhariwal2021diffusion} & GEN & *900  & \textbf{60.3} & \textbf{21.8}\\

\bottomrule
\end{tabular}}
\vspace{-2mm}
\caption{
\textbf{\footnotesize In-domain pre-training on BDD100k.} We follow the recommendation of \cite{el2021large} to pre-train contrastive and masking based self-supervised method with long schedule for small dataset like BDD100k with 70k train images. For semantic segmentation, we use UperNet\cite{xiao2018unified} following official implementation from \cite{yu2020bdd100k}. Reported number is mean IoU at single sacle. For panoptic segmentation, we use panoptic FPN\cite{kirillov2019panoptic}, fine-tune for 36(3$\times$) epochs, reported number is Panoptic Quality (PQ).
}
\label{bdd_indomain_pan}

\end{table}
\begin{table}[t!]
\centering
\resizebox{1.0\linewidth }{!}{ 
\setlength{\tabcolsep}{2pt}
\renewcommand{\arraystretch}{1.0}
\begin{tabular}{lcc|cccc}
\toprule
\multirow{2}{*}{Pre-training (RN-50)}  & PT & Eff. & \multirow{2}{*}{1(2,975)} & \multirow{2}{*}{1/4(744)} & \multirow{2}{*}{1/8(372)} & \multirow{2}{*}{1/30(100)}\\
&task &epoch &&& \\
\midrule
IN sup init. & - & - & 78.7 & 71.3& 65.4& 54.4\\
\midrule
from stratch & - & - & 70.9& 41.8& 40.7& 36.7\\


BYOL\cite{grill2020bootstrap}  & CL & 5000& 73.7 & 54.3 & 49.9 & 44.4 \\ 
Spark\cite{tian2023designing}  & MIM & 2500& 75.7 & 61.2 & 56.0 & 45.3 \\ 

\rowcolor{LightBlue}
DT-feat.distil (ADM) & GEN & *900 & \textbf{77.5} & \textbf{63.9} & \textbf{60.0} & \textbf{50.0} \\ 
\bottomrule

\end{tabular}}
\vspace{-2.5mm}
\caption{
\textbf{\footnotesize Transfer learning: BDD100K to Cityscapes semantic segmentation task.} We pre-trained baselines including ours with unlabeled BDD100K images, and finetuned the backbone on Cityscapes at varying number of labels. Here, we show our method learned transferable features, and achieves the best results in all data portions. Reported numbers are mean IoU, note that effective epochs of our method includes 300 epochs generative model pre-training and 600 epochs feature distillation pre-training. 
} 
\label{bdd_transfer}
\vspace{-1mm}
\end{table}

\begin{table}[!t]
\centering
\resizebox{1.0\linewidth }{!}{ 
\setlength{\tabcolsep}{4pt}
\renewcommand{\arraystretch}{1.0}
\begin{tabular}{lcc|c}
\toprule
Pre-training (ResNet-50) & FID (IN1k) &  Pre. Data & ADE20k(mIoU) \\

\midrule

DT-feat.distil. w/ BigGAN   & 25.3 & Synthetic &40.8 \\
DT-feat.distil. w/ ICGAN  & 17.0 &  Synthetic & 41.2 \\
DT-feat.distil. w/ ADM & 26.2 &   Real & 42.5 \\

\bottomrule
\end{tabular}}
\vspace{-3mm}
\caption{
{\small Ablation study with different unconditional generative models on ImageNet using DreamTeacher.} 
} 
\vspace{-2mm}
\label{ablate_gen_fid}
\end{table}

\begin{table*}[t]
\centering
\resizebox{0.75\textwidth }{!}{ 
\setlength{\tabcolsep}{18pt}
\renewcommand{\arraystretch}{1.0}
\begin{tabular}{lcccc}
\hline
& LSUN & FFHQ & ImagetNet & BDD100K \\
\hline
resolution & 256x256 & 256x256 & 256x256 & 128x256 \\
diffusion steps & 1000 & 1000 & 1000 & 1000 \\
noise Schedule & linear & linear & linear & linear \\
channels & 256 & 256 & 256 & 128 \\
depth & 2 & 2 & 2 & 2 \\
channels multiple & 1,1,2,2,4,4 & 1,1,2,2,4,4 & 1,1,2,2,4,4 & 1,1,2,2,3,4 \\
heads Channels & 64 & 64 & 64 & 32 \\
attention resolution & 32,16,8 & 32,16,8 & 32,16,8 & 16,8 \\
dropout & 0.1 & 0.1 & 0.0 & 0.0 \\
batch size & 256 & 256 & 256 & 256 \\
learning rate & 1e-4 & 1e-4 & 1e-4 & 1e-4 \\
\hline
\end{tabular}}

\caption{
\textbf{Hyperparameters for diffusion models used in the paper.}
} 
\label{diffusion-hyper}
\end{table*}

\section{Generative Models Analysis}
Here we include additional analysis on the effect of using different generative models with DreamTeacher. In Table~\ref{ablate_gen_fid}, we show
DreamTeacher with different generative models on ImageNet. For GAN-based models, ICGAN has better generative
modelling performance compared to BigGAN in terms of
FID (17.0 and 25.3). The backbone pre-trained with ICGAN
also has better transfer learning performance in ADE20K
(41.2 and 40.8). This observation is in line with the empirical results shown in BigBiGAN paper, which found that
generative models with lower FID obtain higher ImageNet
classification accuracy. For GANs, we use synthesized data
to pre-train the image backbone. For diffusion models, we
use encoded data (real images, see Sec.~\ref{sec:unsupreps} main paper),
which makes FID scores of generated data less informative.
In fact, DreamTeacher with diffusion models performs better
in transfer learning, despite ADM’s lower generation FID.

\section{Architecture}
Here we provide additional implementation details about our method's architecture.
\subsection{Implementation: Feature Regressors}
Our Feature Regressor is implemented as a Feature Pyramid Network (FPN)\cite{lin2017feature} with additional Pyramid Pooling Module (PPM)\cite{zhao2017pyramid}. We use pool scale at 1,2,3,6. We add batch normalization and ReLU activation for both the lateral connection and the bottom-up convolution blocks in FPN. We use feature channel size 256. We also add 1x1 conv layer at the end of each level to map the output feature channel to the generator's feature channel.
\subsection{Implementation: Feature Interpreter}
Feature Interpreter is implemented as a series of Feature Fuse layer to map and fuse generator's features into a logit map. Feature Fuse Layer is implemented by a block of 1x1 conv, bilinear upsampling, concatenation, and Depth-wise Separable Convolution\cite{chollet2017xception}. We applied Group Norm\cite{wu2018group} with group number 32 and Swish Activation \cite{ramachandran2017searching} in-between blocks. The feature dimension is 256. We also apply dropout with rate 0.1 before the 1x1 conv mapping to the output logits.

\section{ImageNet Benchmarks}
\subsection{Implementation: generative models}
For GAN based generative models, we use pre-trained unconditional BigGAN~\cite{brock2018large} from this repository\footnote{https://github.com/lukemelas/pytorch-pretrained-gans},
ICGAN\cite{casanova2021instance} from this repository\footnote{https://github.com/facebookresearch/ic\_gan}. For diffusion based generative model, we use ADM without classifier guidance, pre-trained with resolution 256x256 from this repository \footnote{https://github.com/openai/guided-diffusion}. Please also refer to Table~\ref{diffusion-hyper} for the hyperparameters used in training the diffusion model on ImageNet.

\subsection{Implementation: pre-training}
We pre-trained ResNet-50 backbone by diffusing images for 150 steps and running a single denoising step to extract the ADM's UNet decoder features at blocks 3,6,9, and 12. We use the LAMB~\cite{you2019large} optimizer with batch size of 2048 and $lr=4e-3$ with cosine decay learning rate schedule and pre-trained for 200 epochs on the ImageNet training set without class labels. We center crop images into resolution 256x256 and only apply horizontal flipping as the data augmentation. Please see Table~\ref{pre-train-hyper} for other hyperparameters.

\subsection{Implementation: downstream tasks}

\nbf{ImageNet: classification fine-tuning} For ResNet50, we follow \cite{tian2023designing} to use the latest open-source ResNet A2 schedule from \cite{wightman2021resnet}. For ConvNeXt-B, we use the official implementation. We did not tune the hyper-parameters from \cite{tian2023designing}. Please refer to SparK official github repo\footnote{https://github.com/keyu-tian/SparK/} for the hyper-parameters.

\nbf{ADE20K: semantic segmentation}
We use UperNet implemented in MMSegmentation for training ADE20K semantic segmentation task. We use the default hyperparameters from MMSegmentation and train for 160K iterations. We use the same hyperparameters for both the baselines and our methods. After finetuning, we evaluate the performance on ADE20K validation set without using multi-scale flip augmentation.

\nbf{MSCOCO: instance segmentation}
For ResNet-50, we follow \cite{tian2023designing} to use Mask R-CNN with R50-FPN backbone implemented in Detectron2\footnote{https://github.com/facebookresearch/detectron2} for MSCOCO instance segmentation task. We use the same configuration as in SparK\cite{tian2023designing} to finetune the pre-trained backbone for 12 epochs (1$\times$ schedule) or 24 epochs (2$\times$) and evaluate performance on the MSCOCO 2017 validation set. For ConvNeXt-B, we follow the configuration of iBOT\cite{zhou2021ibot} to use Cascade Mask R-CNN, fine-tune for 12 epochs. Following the convention, we do not use multi-scale testing, large-scale jittering augmentation, or soft-NMS in all our COCO experiments.

\nbf{BDD100K: instance segmentation}
We use Mask R-CNN with R50-FPN backbone implemented in MMDetection for BDD100K instance segmentation task. We use the official data processing script from this repository\footnote{https://github.com/SysCV/bdd100k-models/tree/main/ins\_seg} and finetuned pre-trained backbone with 36 epochs (3x schedule) and evaluate performance on BDD100K validation set.

\section{BDD100K Benchmarks}
\subsection{Implementation: Generative Models}
\nbf{StyleGAN2} We resize BDD100K images from the original resolution 720x1280 to 512x1024 when training the GAN. We use the default configuration from~\cite{karras2020analyzing} without adaptive augmentation. We empirically found that turning off path length regularization and style mixing loss improves data sample quality on the BDD100k driving scenes. We train the network with batch size 32 and $gamma=10.0$ until convergence.

\nbf{Diffusion Models}  
We train our own diffusion model on BDD100K, and we summarize training hyperparameters in table~\ref{diffusion-hyper}. We resize BDD100K images from the original resolution 720x1280 to 128x256 resolution. We use diffusion model architecture builds on the U-Net by ~\cite{dhariwal2021diffusion}, and we use linear noise schedule from $1e-4$ to $2e-2$. 
\begin{table*}[t!]
\centering
\resizebox{0.9\textwidth }{!}{ 
\setlength{\tabcolsep}{14pt}
\renewcommand{\arraystretch}{1.0}
\begin{tabular}{lcccc}
\hline
& LSUN & FFHQ & ImagetNet & BDD100K \\
\hline
resolution & 256x256 & 256x256 & 256x256 & 128x256/512x1024 \\
noise steps & 50 & 50 & 150 & 50 \\
optimizer & AdamW & AdamW & LAMB & AdamW \\
base learning rate & 4e-3 & 4e-3 & 4e-3 & 4e-3 \\
weight decay & 0.05 & 0.05 & 0.05 & 0.05 \\
optimizer momentum & $\beta_1,\beta_2=0.9,0.95$ & $\beta_1,\beta_2=0.9,0.95$ & $\beta_1,\beta_2=0.9,0.95$ & $\beta_1,\beta_2=0.9,0.95$ \\
batch size & 256 & 256 & 2048 & 256 \\
training epochs & 100 & 100 & 200 & 600 \\
learning rate schedule & cosine decay & cosine decay & cosine decay & cosine decay \\
warmup epochs & 20 & 20 & 30 & 30 \\
augmentation & horizontal flip & horizontal flip & horizontal flip & horizontal flip \\
\hline
\end{tabular}}

\caption{
\textbf{Hyperparameters for pre-training the image backbones.}
} 
\label{pre-train-hyper}
\end{table*}

\subsection{Implementation: Pre-training}
When DreamTeacher pre-training using the StyleGAN2 generator, we use resolution 512x1024. We use resolution 128x256 for the ADM generator, same resolution as in the generative model training. We use AdamW\cite{loshchilov2017decoupled} optimizer with learning rate $lr=4e-3$ and cosine decay learning schedule to train 600 epochs. Note that we only use horizontal flip as the augmentation during pre-training. Please see Table~\ref{pre-train-hyper} for information about the hyperparameters.
\subsection{Implementation: Downstream Tasks}
For all three tasks, we use the official data split and dataset configuration from the official BDD100K repository \footnote{https://github.com/SysCV/bdd100k-models}.

\nbf{Semantic segmentation}
We use UperNet~\cite{xiao2018unified} implemented in MMSegmentation~\footnote{https://github.com/open-mmlab/mmsegmentation} for evaluating the semantic segmentation task on BDD100K. We train UperNet with the pre-trained ResNet-50 backbone using SGD optimizer with $lr=0.01$, momentum=0.9, weight decay=0.0005, and poly learning rate schedule with power=0.9 for 80k iterations. We use the same training hyperparameters for both the baseline models and our models. After finetuning, we evaluate the model on BDD100K validation set without multi-scale flip test-time augmentation.

\nbf{Instance segmentation}
We use Mask R-CNN~\cite{he2017mask} with R50-FPN backbone implemented in MMDetection\footnote{https://github.com/open-mmlab/mmdetection} for evaluating the instance segmentation task on BDD100K. We only use the pre-trained weights of ResNet-50 during finetuning and the rest of the networks are randomly initialized. We use SGD optimizer with $lr=0.02$, momentum=0.9, and weight decay=0.0001 to train Mask R-CNN for 36 epochs (3x schedule). We use the same training hyperparameters for both the baseline models and our models.

\nbf{Panoptic segmentation}
We use PanopticFPN~\cite{kirillov2019panoptic} with R50-FPN backbone implemented in MMDetection for evaluating panoptic segmentation task on BDD100K. We only use the pre-trained weights of ResNet-50 during finetuning and the rest of the networks are randomly initialized. We use the SGD optimizer with $lr=0.02$, momentum=0.9, and weight decay=0.0001 to train PanopticFPN for 36 epochs (3x schedule). We use the same training hyperparameters for both the baseline models and our models.

\nbf{Cityscapes transfer learning} In this experiment, both the baselines and our method use ResNet-50 backbone pre-trained on 70k unlabeled images from the BDD100K training set. After pre-training, we finetuned the backbone with UPerNet~\cite{xiao2018unified} for semantic segmentation tasks in Cityscapes. We follow the public split~\cite{chen2021semi} to split 2,975 training images into 1/4, 1/8, and 1/30 subsets. We use MMSegmentation to train UperNet with 80k iterations for the full set and 20k iterations for the subsets. We use the default hyperparameters for all the baselines and our method. After finetuning, the model is evaluated on the official validation set (5,000 images).

\section{Label-Efficient Benchmarks}
\subsection{Implementation: generative models}
For LSUN cat, horse and bedroom dataset, we use pre-trained ADM models from the guided-diffusion models' repository\footnote{https://github.com/openai/guided-diffusion}. For FFHQ, we use pre-trained model from this repository \footnote{https://github.com/yandex-research/ddpm-segmentation}, following DDPM-seg~\cite{henaff2021efficient}. Please see Table~\ref{diffusion-hyper} for the hyperparmeters used in training the diffusion models.

\subsection{Implementation: Feature Interpreter} We train the feature interpreter by first diffusing the real images with 50 time steps, and then extract the ADM's features by running one step of denoising. We use ADM's UNet decoder features at block 3,6,9, and 12. We then train the feature interpreter branch with AdamW optimizer with $lr=4e-3$, weight decay $0.05$, $\beta_1, \beta_2=0.9, 0.95$, warm-up epochs 20 and train for 100 epochs. We only use horizontal flip when training the feature interpreter.

\subsection{Implementation: pre-training}
We use the AdamW optimizer with learning rate $lr=4e-3$ and cosine decay learning schedule to train 100 epochs. Please see Table~\ref{pre-train-hyper} for hyperparmeters for different datasets. Note that we only use horizontal flipping as data augmentation during pre-training.

\subsection{Implementation: downstream tasks}
We use UperNet implemented in MMSegmentation for semantic segmentation tasks. We use the default hyperparameters in MMSegmentation. For all four tasks, we train UperNet using 20k iteration schedule. For feature distillation, only the backbone is initialized from pre-trained weight, and the rest are randomly initialized. For mix-distillation, we initialize the backbone as well as the UperNet with our pre-trained weights.

\newcommand\hhh{2.2cm}
\newcommand\www{4.0cm}

\begin{figure*}[h!]
\vspace{-3mm}
\centering
\setlength{\tabcolsep}{1pt}
\resizebox{1\linewidth }{!}{ 
\begin{tabular}{ccccc}
\rot{\scriptsize{Instance Seg.}} & \includegraphics[height=\hhh,width=\www, trim=0 0 0 0,clip]{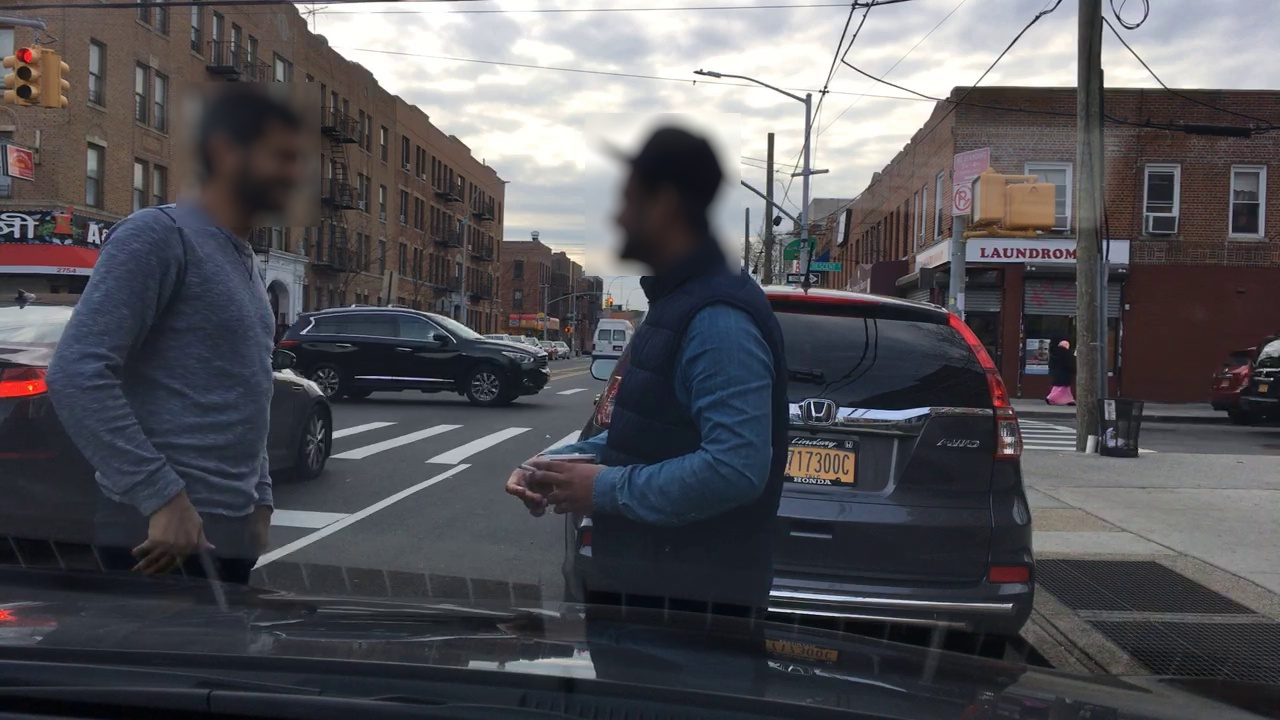} & \includegraphics[height=\hhh,width=\www, trim=0 0 0 0,clip]{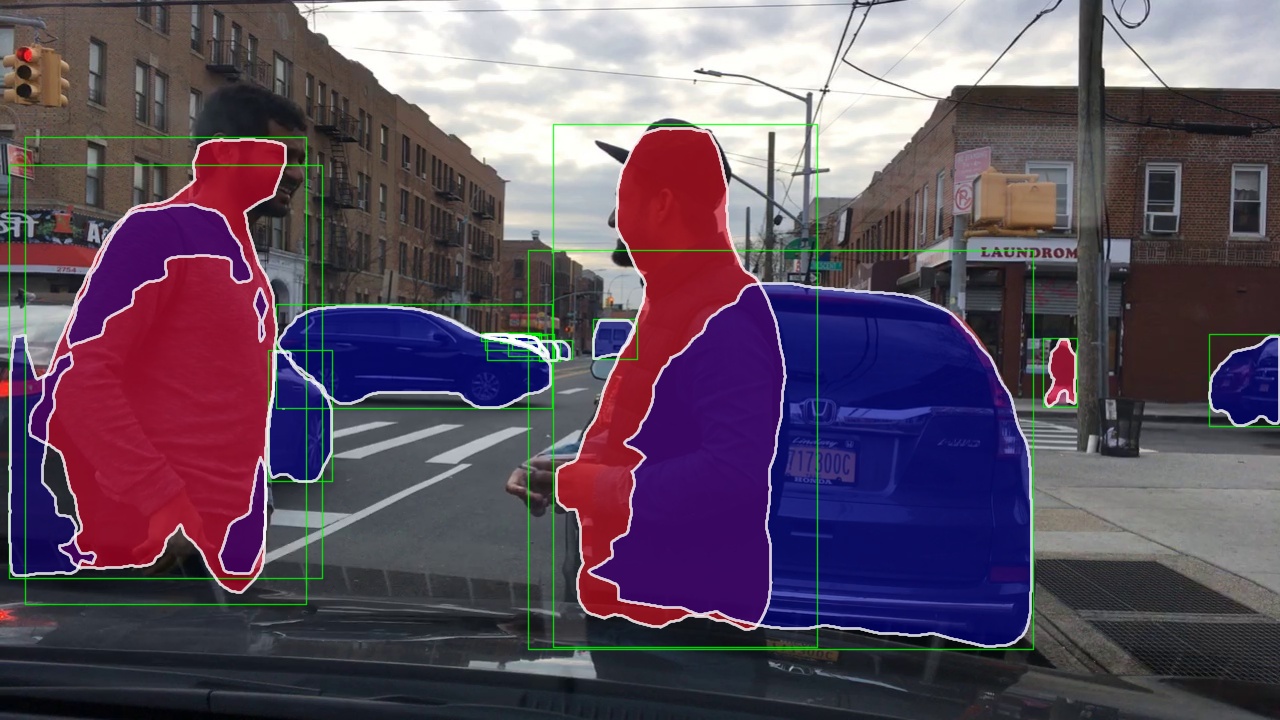} & 
\includegraphics[height=\hhh,width=\www, trim=0 0 0 0,clip]{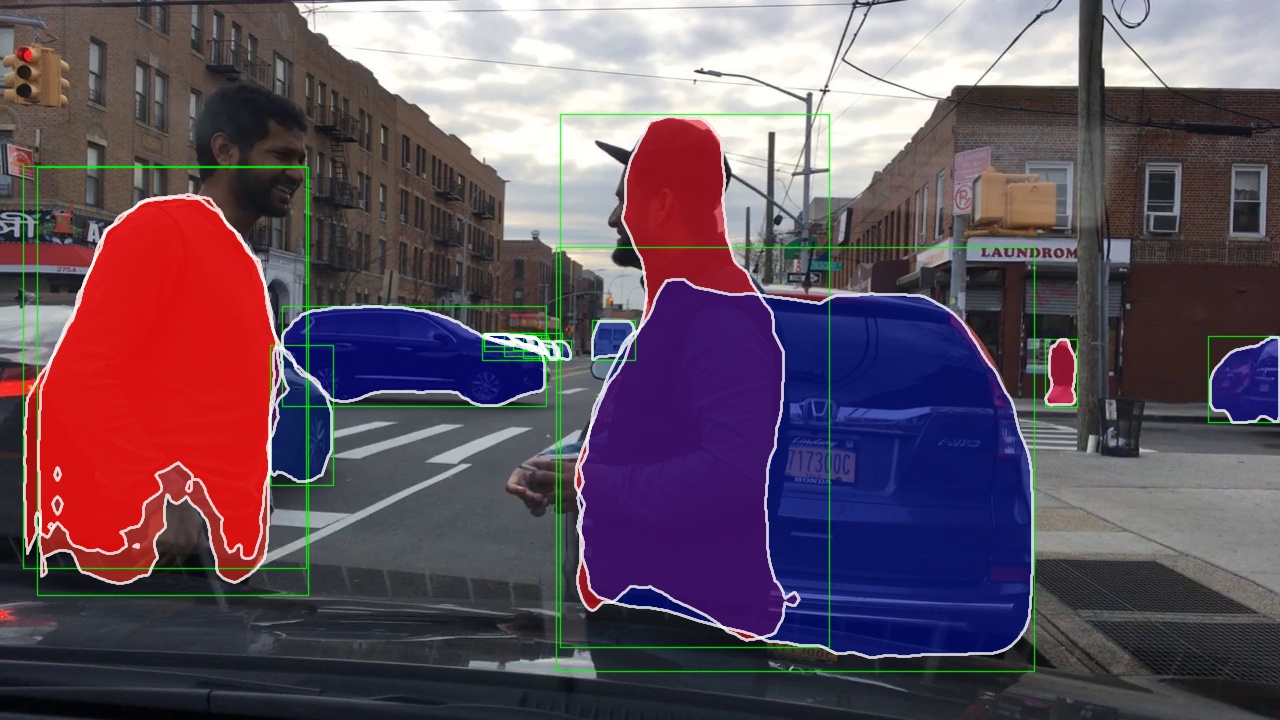} &
\includegraphics[height=\hhh,width=\www, trim=0 0 0 0,clip]{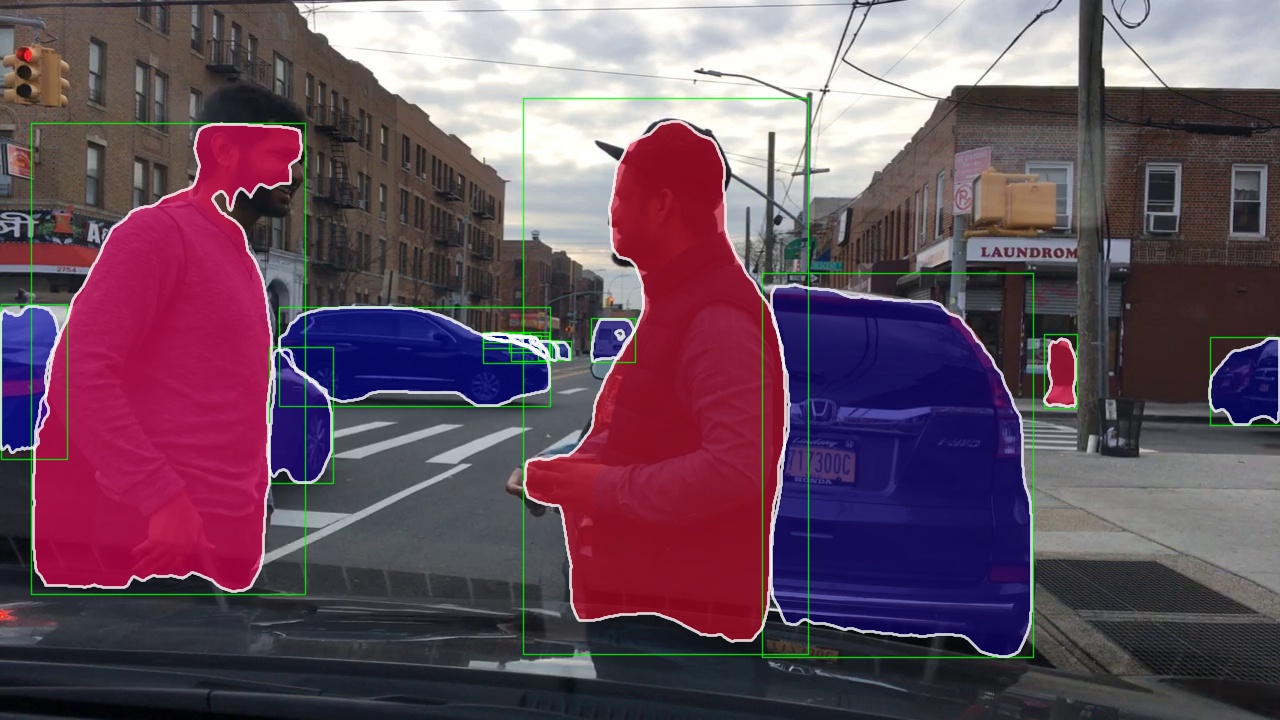} \\

\rot{\scriptsize{Semantic Seg.}}&\includegraphics[height=\hhh,width=\www, trim=0 0 0 0,clip]{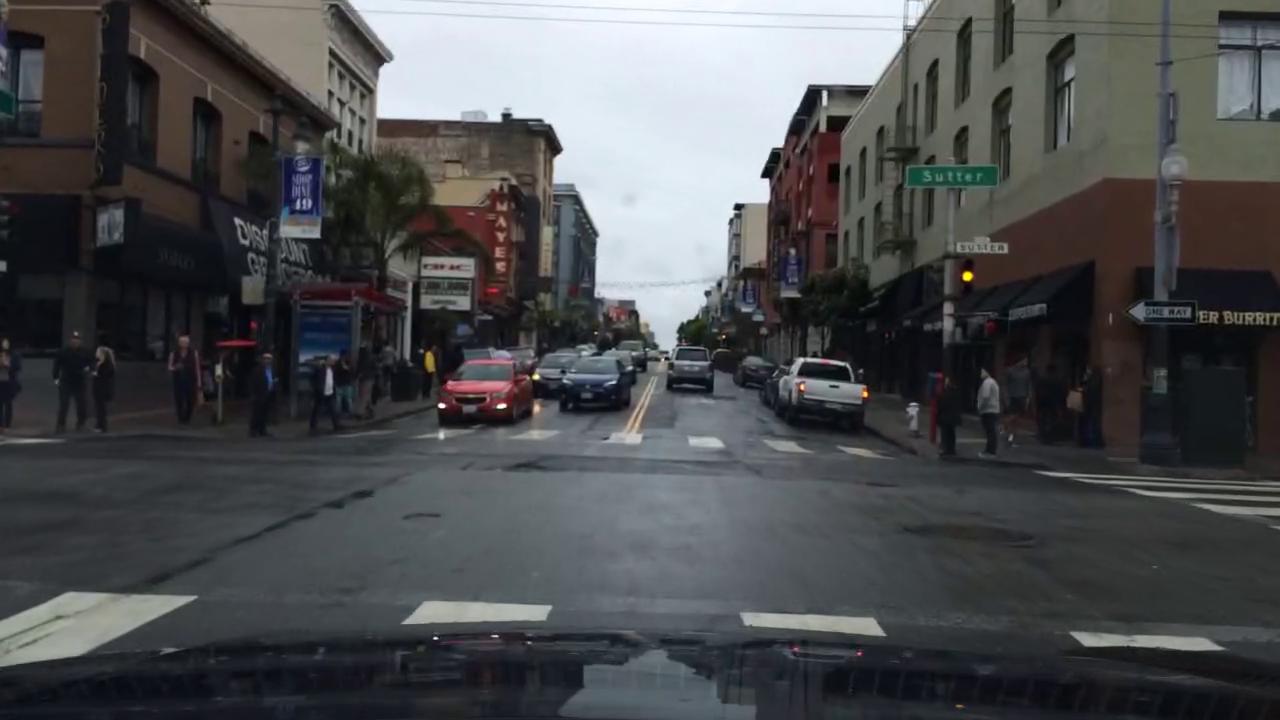} &
\includegraphics[height=\hhh,width=\www, trim=0 0 0 0,clip]{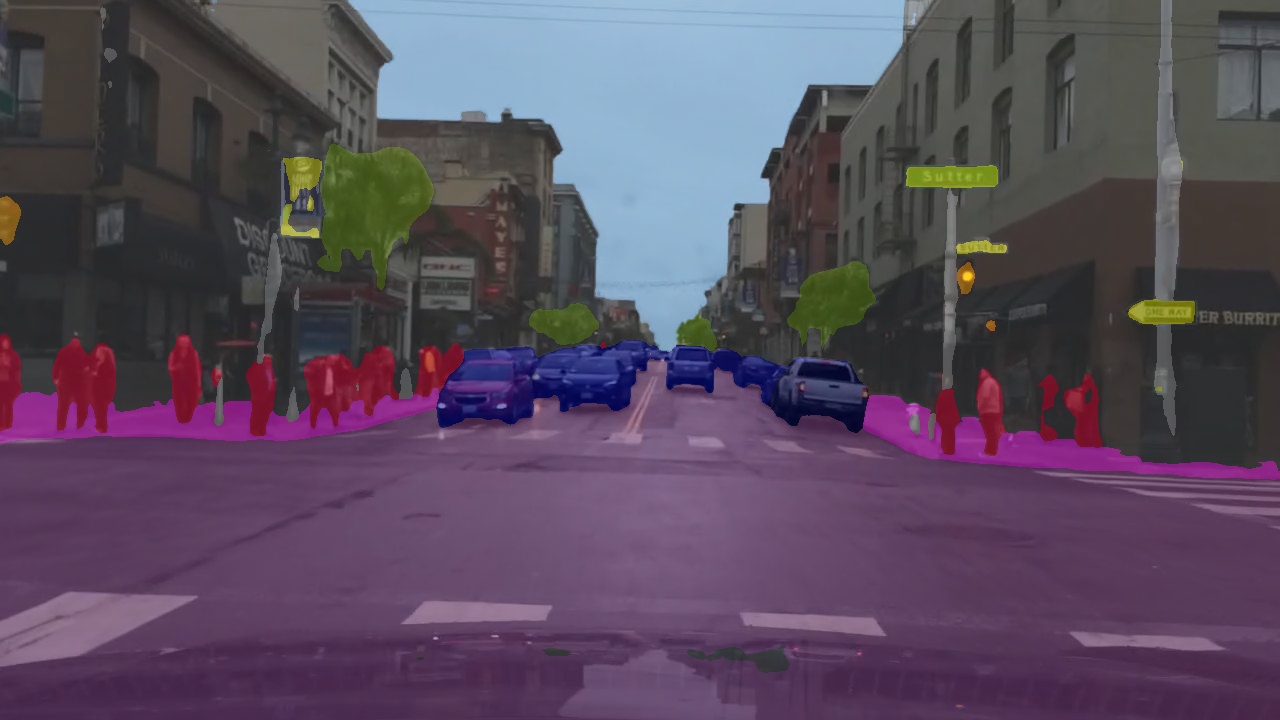} & 
\includegraphics[height=\hhh,width=\www, trim=0 0 0 0,clip]{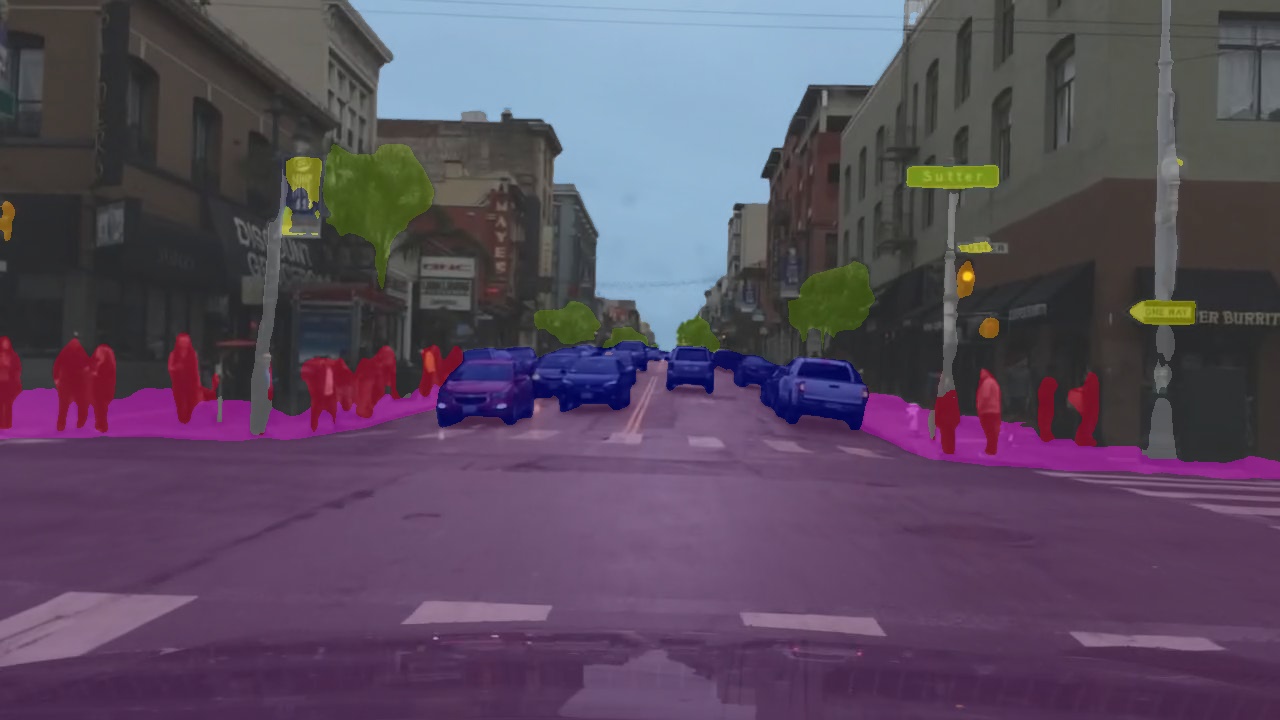} &
\includegraphics[height=\hhh,width=\www, trim=0 0 0 0,clip]{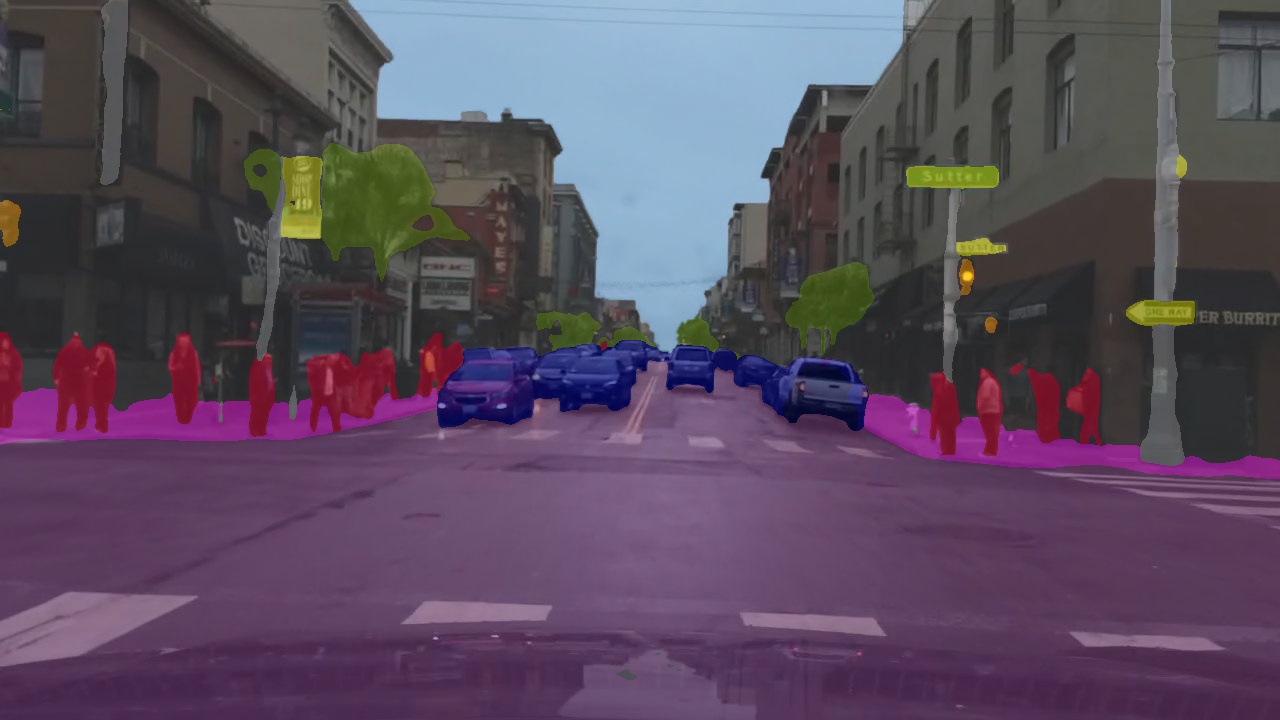} \\

&\scriptsize{Image} &  \scriptsize{denseCL IN-1k-1M pre-trained} &  \scriptsize{DT-feat.distil. BDD100K pre-trained } &  \scriptsize{DT-feat.distil. IN-1k-1M pre-trained }  \\

\end{tabular}}
\hfill
\vspace{-3mm}
\caption{\footnotesize\textbf{Qualitative results on BDD100k Inst./Sem. Seg.} 
Compared with denseCL, our method pre-trained on ImagetNet predicts the correct box on pedestrians and occluded cars, and the mask boundaries are clearer.  On semantic segmentation (second row), our prediction segments traffic signs and thin objects like poles. We blur pedestrian faces in the figure, while the methods make predictions on original images. 
}
\label{fig:bdd100k-results}
\vspace{-2mm}
\end{figure*}

\section{Visualization}
We first provide qualitative comparison of instance/sementic segmentation in BDD100K with baseline and then show visualization of ADM denoising network features at different resolution of driving scenes in BDD100K and feature activation maps of a pre-trained ResNet50 backbone using DreamTeacher on ImageNet1k. We then show semantic segmentation results on FFHQ, LSUN-cat, horse and bedroom. The backbone is ConvNX-b pre-trained with our mix-distillation method. We also show semantic, instance and panoptic segmentation results of our method on BDD100K. Note that results of all three tasks are pre-trained using our DreamTeacher feature distillation method on ResNet-50 backbone. We show results on pre-training on BDD100K in-domain data solely and ImageNet-1k-1M as general domain data. 

\subsection{Qualitative Comparison}
In Figure~\ref{fig:bdd100k-results}, we show qualitative results on BDD100k instance/semantic segmentation task. Comparing to denseCL, our DreamTeacher pre-training method performs better at small/thin objects like traffic signs and poles when fine-tuned on downstream tasks.

\subsection{Feature Activation Maps}
In Figure~\ref{fig:ddpm-steps-feat}, we show feature activation maps at different resolution blocks and different noise steps from ADM~\cite{dhariwal2021diffusion} pre-trained on BDD100k without classifier guidance. We visualize multi-scale features at different resolution blocks, showing features in lower resolution focus on structures, and focus on parts in higher resolution. In Figure~\ref{fig:backbone-feature}, we show feature activation maps on ResNet50 backbone pre-trained with DreamTeacher, the backbone
learns coarse features at lower level layers and finer features at higher level layers.
\begin{figure}[t!]
\vspace{-2mm}
\begin{center}
\includegraphics[width=1.0\linewidth]{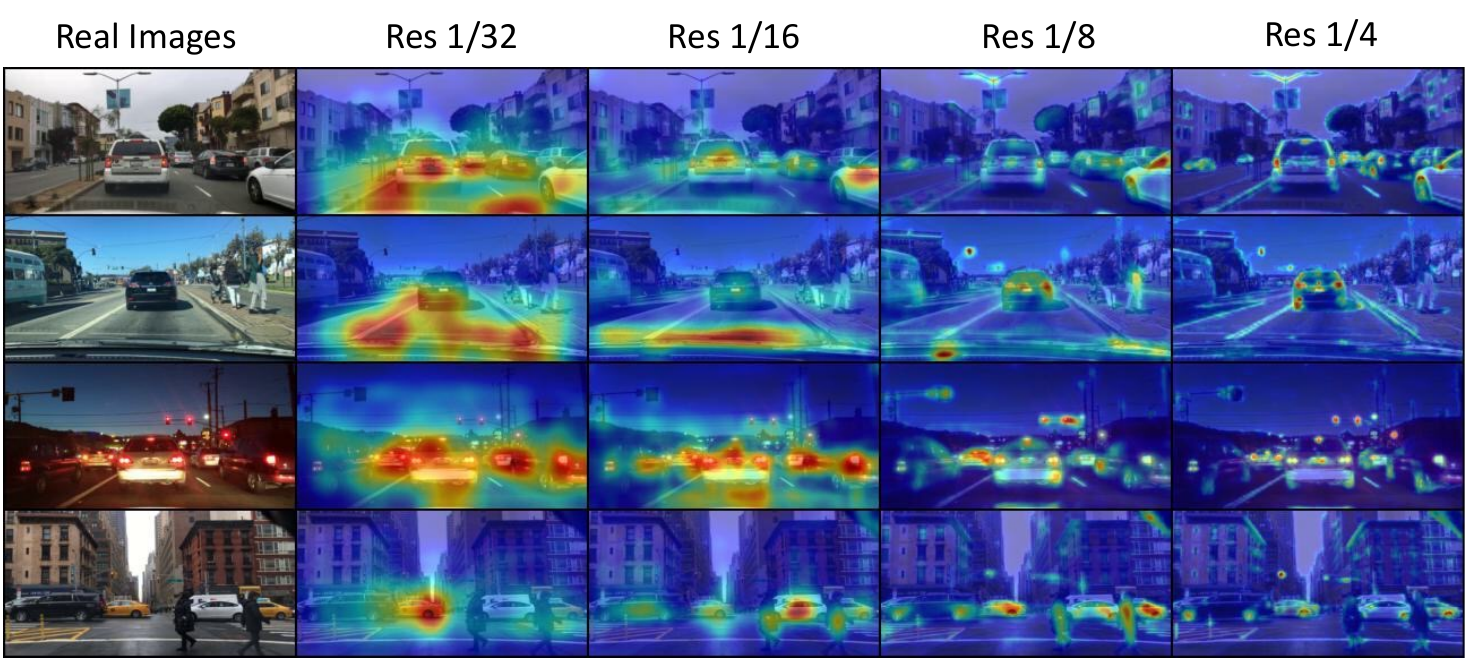}
\end{center}
\vspace{-6mm}
\caption{\textbf{\footnotesize ADM feature visualization (BDD100k)}.
BDD100k image is first diffused by 50 steps and we run one denoising step of the ADM model to extract the feature. We see that features in the low resolution block focus on scene layouts and objects, and in higher resolution, they focus on parts like car wheels, and traffic lights.
}
\label{fig:ddpm-steps-feat}
\vspace{-2mm}
\end{figure}

\begin{figure}[t]
\begin{center}
\includegraphics[width=1.0\linewidth, height=0.4\linewidth]{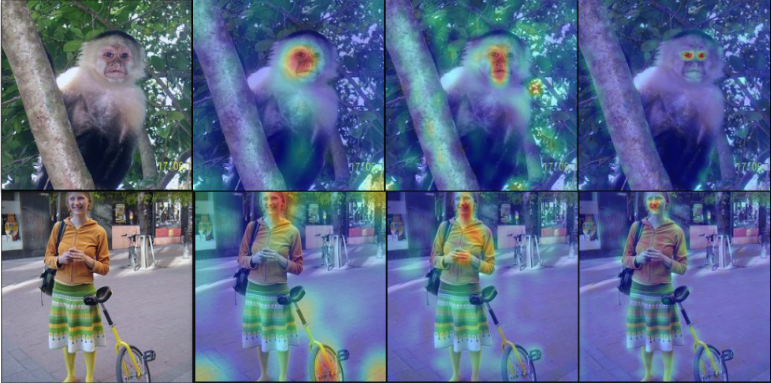}
\end{center}
\vspace{-6mm}
\caption{\small \textbf{\footnotesize DreamTeacher pre-trained ResNet50 backbone feature activation maps on ImageNet images}. From left to right, we show the image and features at 1/32, 1/16, and 1/8 input resolution.
}

\label{fig:backbone-feature}
\vspace{-1mm}
\end{figure}

\subsection{Label-Efficient Semantic Segmentation}
In Figure~\ref{fig:ffhq-seg}, we show semantic segmentation results on FFHQ unlabeled images.
In Figure~\ref{fig:bedroom-seg}, we show semantic segmentation results on LSUN-bedroom unlabeled images.
In Figure~\ref{fig:cat-seg}, we show semantic segmentation results on LSUN-cat unlabeled images.
In Figure~\ref{fig:horse-seg}, we show semantic segmentation results on LSUN-horse unlabeled images.
From the visulization, our method is robust to cat in different pose, multi objects occurs in the same images (horse/cat). 

\subsection{BDD100K: semantic segmentation}
In Figure~\ref{fig:bdd100k-seg-merge-in-domain}, we show semantic segmentation results on BDD100K, pre-trained by DreamTeacher feature distillation on BDD100K unlabeled dataset. And in figure ~\ref{fig:bdd100k-seg-merge-imagenet}, we show results pre-trained on ImageNet unlabeled dataset with our method. Comparing to BDD100K pre-trained, ImageNet pre-trained method works better with rare and small objects like rider and traffic lights.

\subsection{BDD100K: instance segmentation}
In Figure~\ref{fig:bdd100k-ins-merge-in-domain}, we show instance segmentation results on BDD100K, pre-trained by DreamTeacher feature distillation on BDD100K unlabeled dataset. and 
in figure~\ref{fig:bdd100k-ins-merge-imagenet}, we show instance segmentation results on BDD100 on ImageNet unlabeled dataset. As a comparison, model pre-trained on ImageNet detect and segment small objects better.

\subsection{BDD100K: panoptic segmentation}
In Figure~\ref{fig:bdd100k-pan-merge-in-domain}, we show panoptic segmentation results on BDD100K, pre-trained by DreamTeacher feature distillation on BDD100K unlabeled dataset. And in figure~\ref{fig:bdd100k-pan-merge-imagenet}, we show panoptic segmentation results on BDD100K, pre-trained on ImageNet unlabeled dataset. Note that model pre-trained with BDD100K performs well on things class like road and tree etc, but model pre-trained on ImageNet gets clearer boundary, especially for small objects.

\begin{figure*}[t]
\begin{center}
\includegraphics[width=0.95\textwidth]{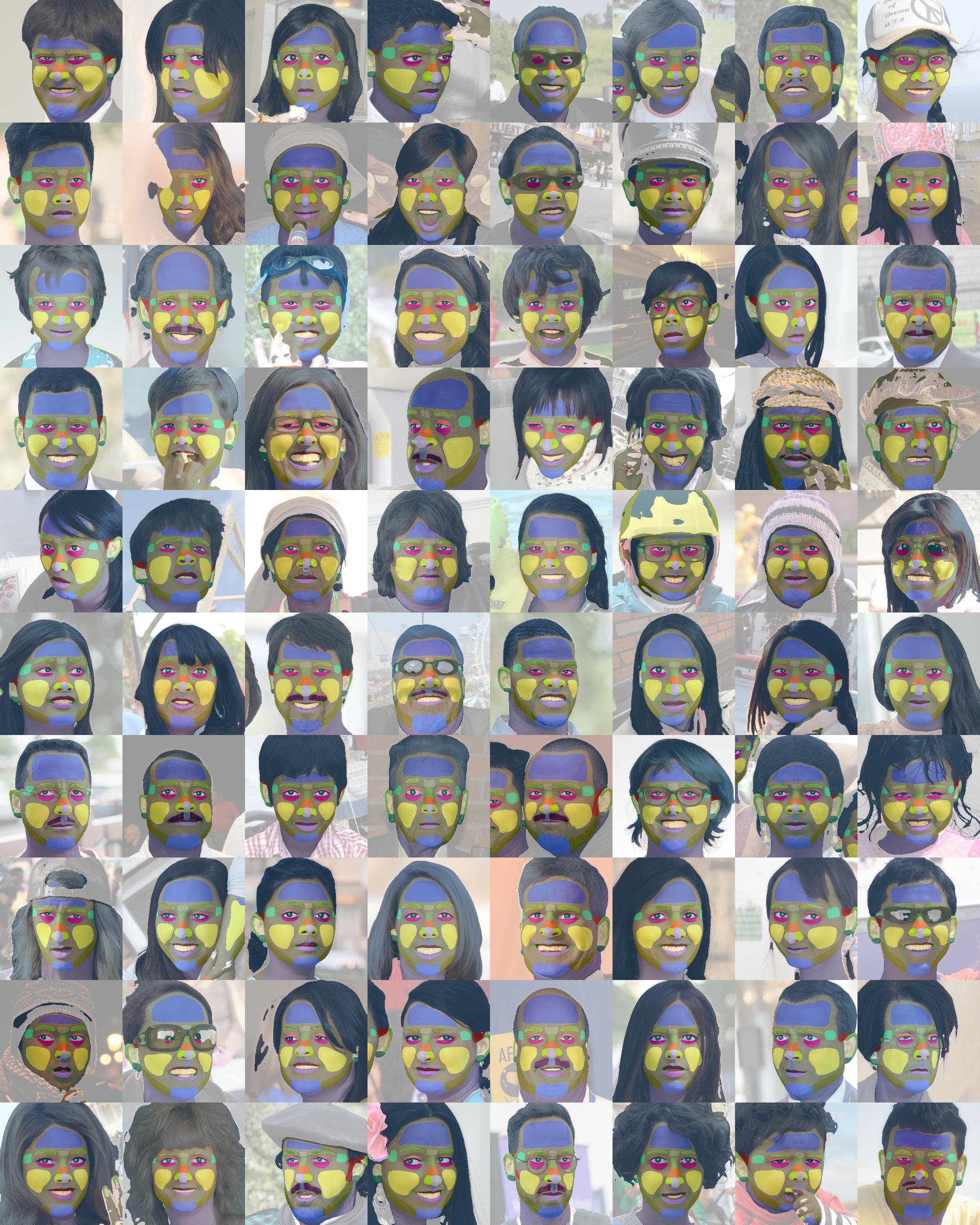}
\end{center}

\caption{\textbf{Semantic segmentation: FFHQ with 34 classes.} Qualitative results of our ConvNX-B model pre-trained with DreamTeacher-feature distillation on FFHQ unlabelled images.
}
\label{fig:ffhq-seg}

\end{figure*}

\begin{figure*}[t]
\begin{center}
\includegraphics[width=0.95\textwidth]{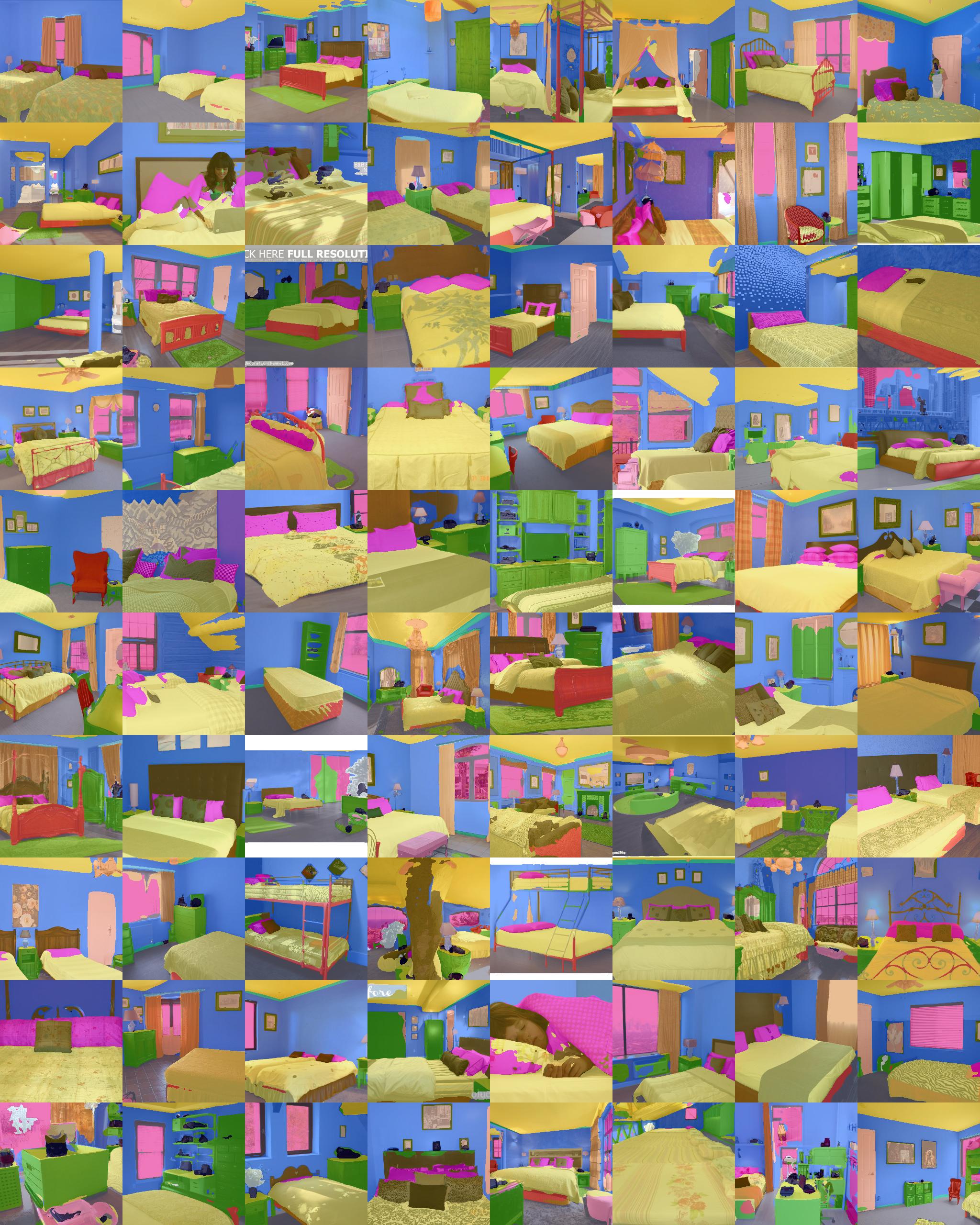}
\end{center}

\caption{\textbf{Semantic segmentation: LSUN-bedroom with 28 classes.} Qualitative results of our ConvNX-B model pre-trained with DreamTeacher-feature distillation on LSUN-bedroom unlabelled images.
}
\label{fig:bedroom-seg}

\end{figure*}

\begin{figure*}[t]
\begin{center}
\includegraphics[width=0.95\textwidth]{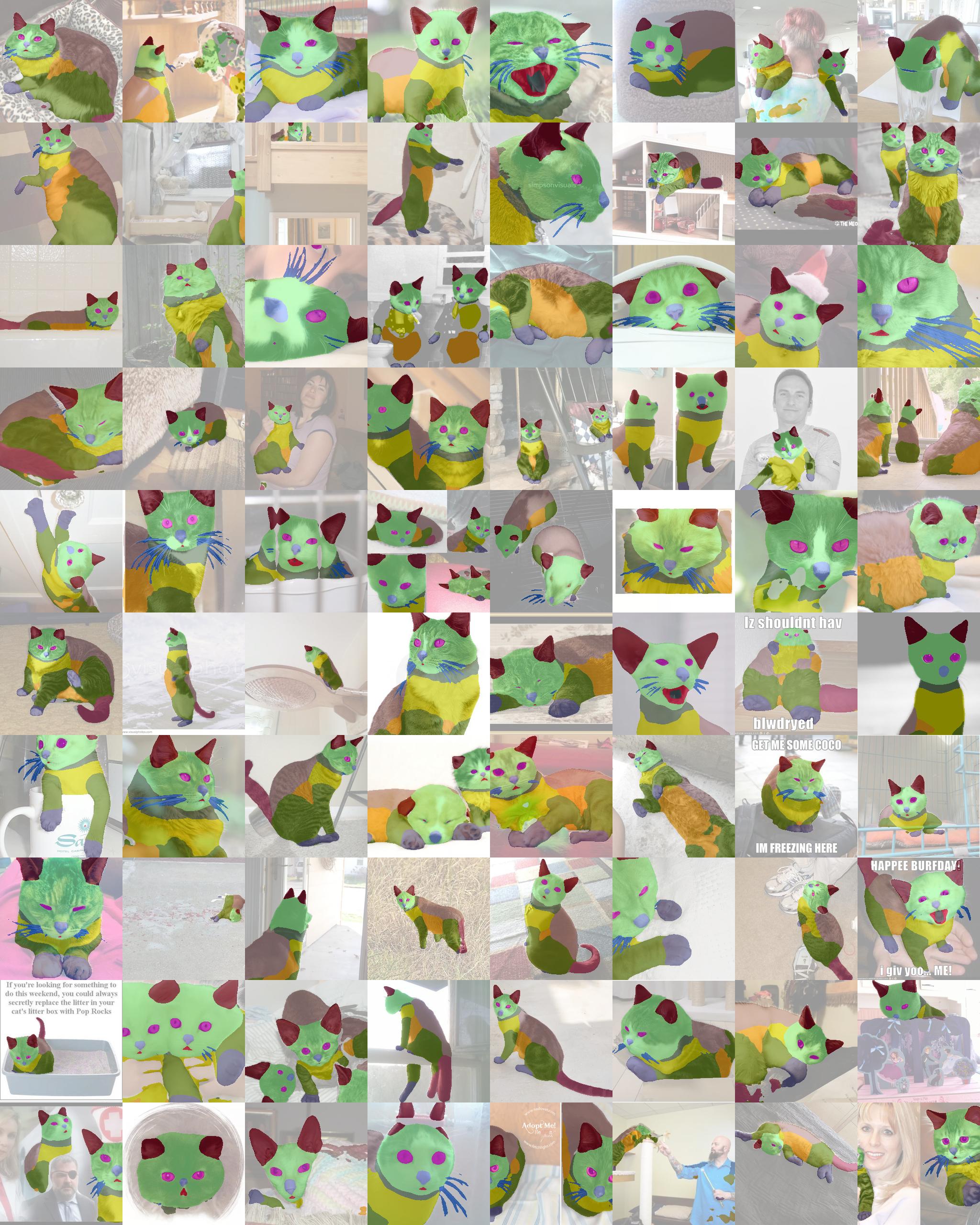}
\end{center}

\caption{\textbf{Semantic segmentation: LSUN-cat with 15 classes.} Qualitative results of our ConvNX-B model pre-trained with DreamTeacher-feature distillation on LSUN-cat unlabelled images.
}
\label{fig:cat-seg}

\end{figure*}

\begin{figure*}[t]
\begin{center}
\includegraphics[width=0.95\textwidth]{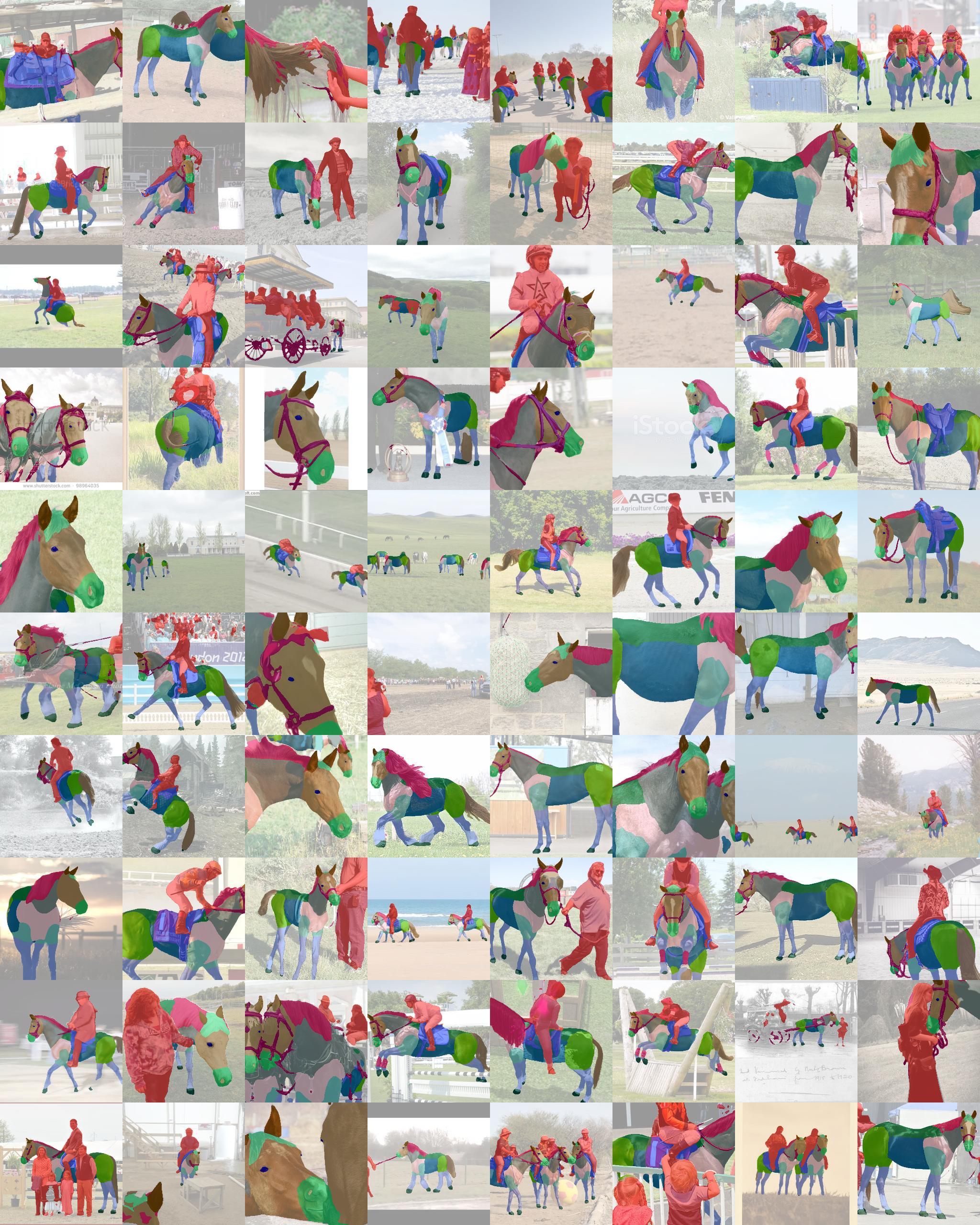}
\end{center}

\caption{\textbf{Semantic segmentation: LSUN-horse with 21 classes.} Qualitative results of our ConvNX-B model pre-trained with DreamTeacher-feature distillation on LSUN-horse unlabelled images.
}
\label{fig:horse-seg}

\end{figure*}

\begin{figure*}[t]
\begin{center}
\includegraphics[width=1.0\textwidth]{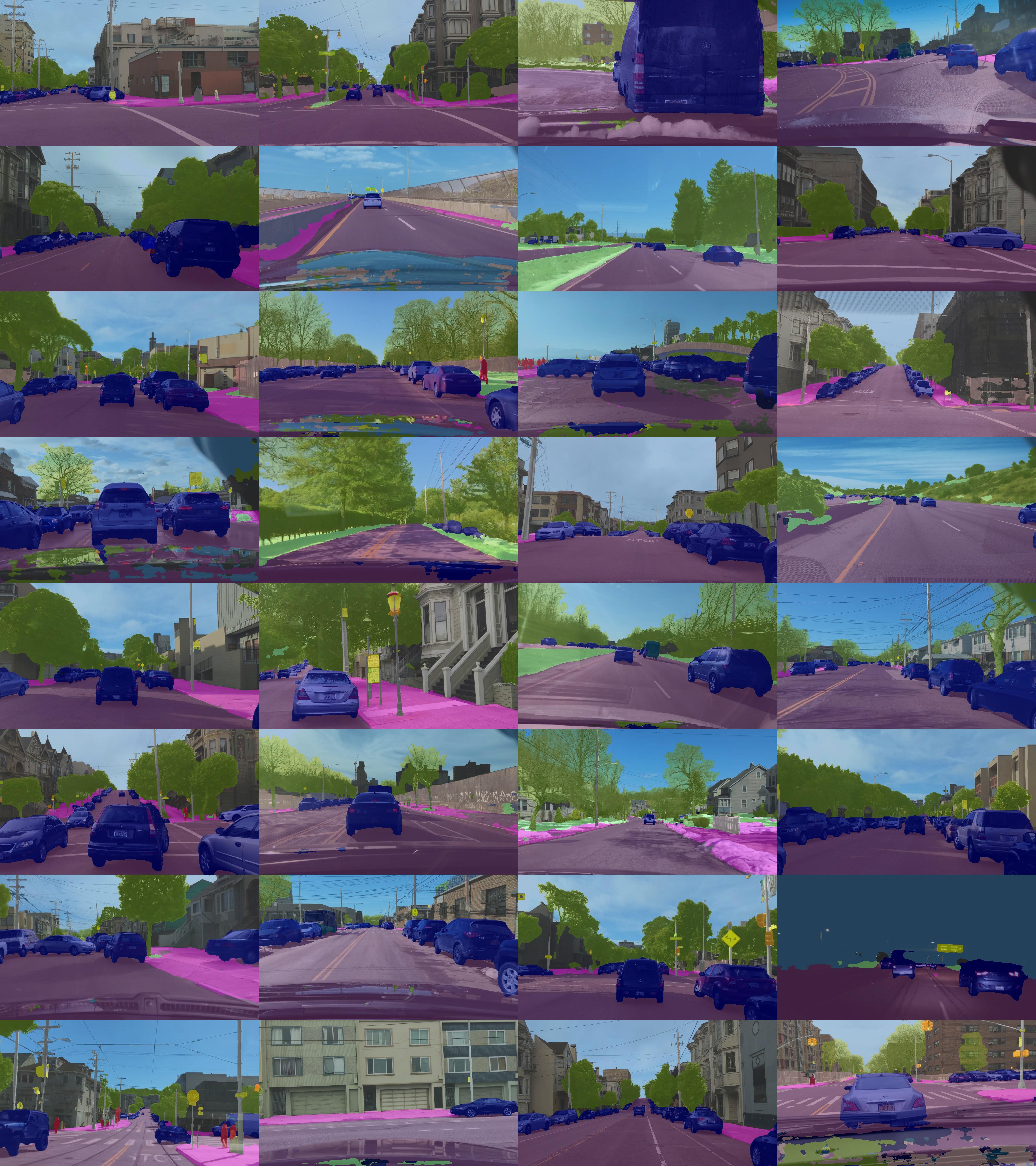}
\end{center}

\caption{\textbf{BDD100K semantic segmentation visualization: pre-trained with DreamTeacher feature distillation on BDD100K.} The backbone is resnet-50, finetuned using UperNet.
}
\label{fig:bdd100k-seg-merge-in-domain}

\end{figure*}

\begin{figure*}[t]
\begin{center}
\includegraphics[width=1.0\textwidth]{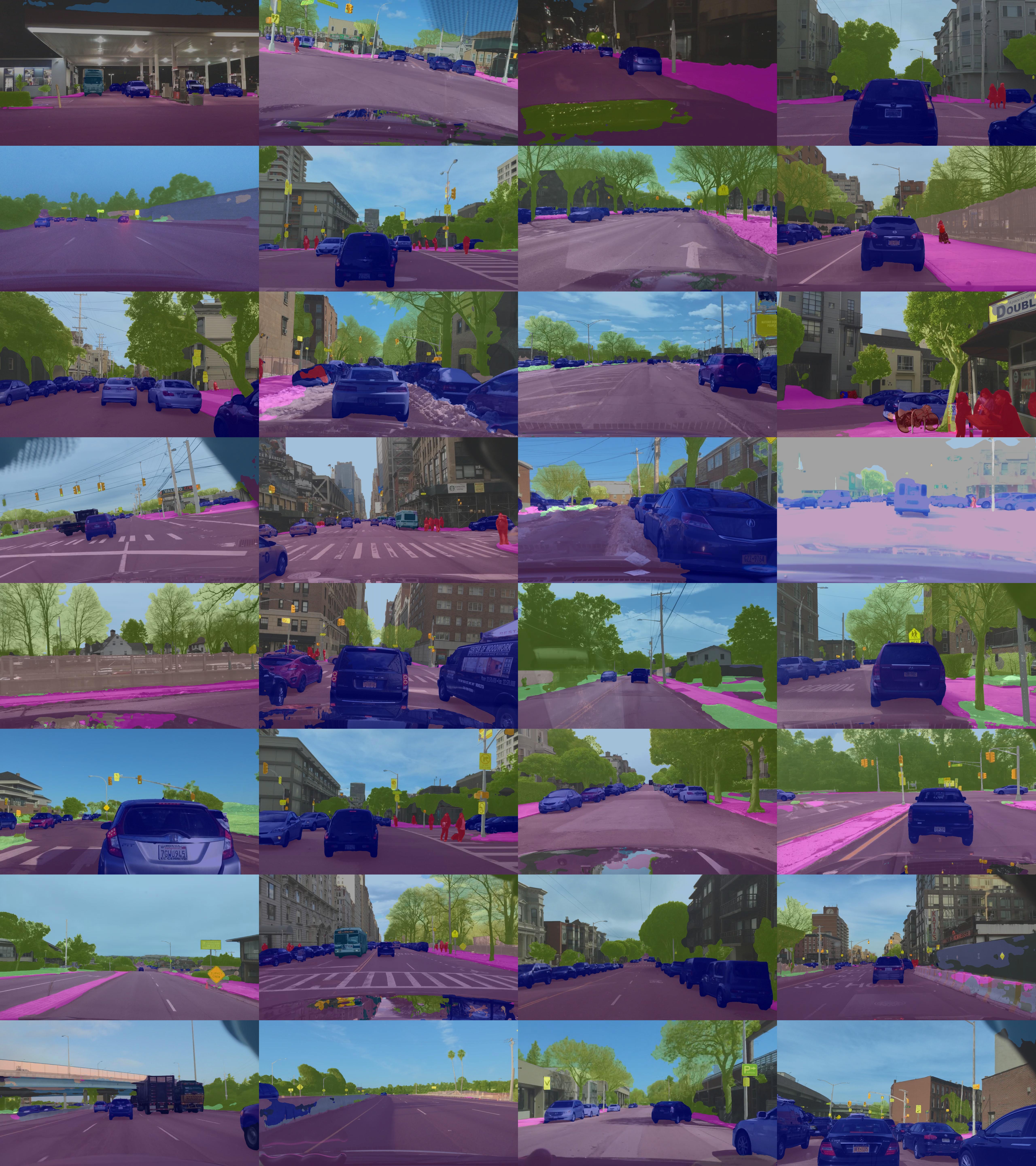}
\end{center}

\caption{\textbf{BDD100K semantic segmentation visualization: pre-trained with DreamTeacher feature distillation on IN1k-1M.} The backbone is resnet-50, finetuned using UperNet. Only the backbone weight is pre-trained, other part of the networks are randomly initialized.
}
\label{fig:bdd100k-seg-merge-imagenet}

\end{figure*}

\begin{figure*}[t]
\begin{center}
\includegraphics[width=1.0\textwidth]{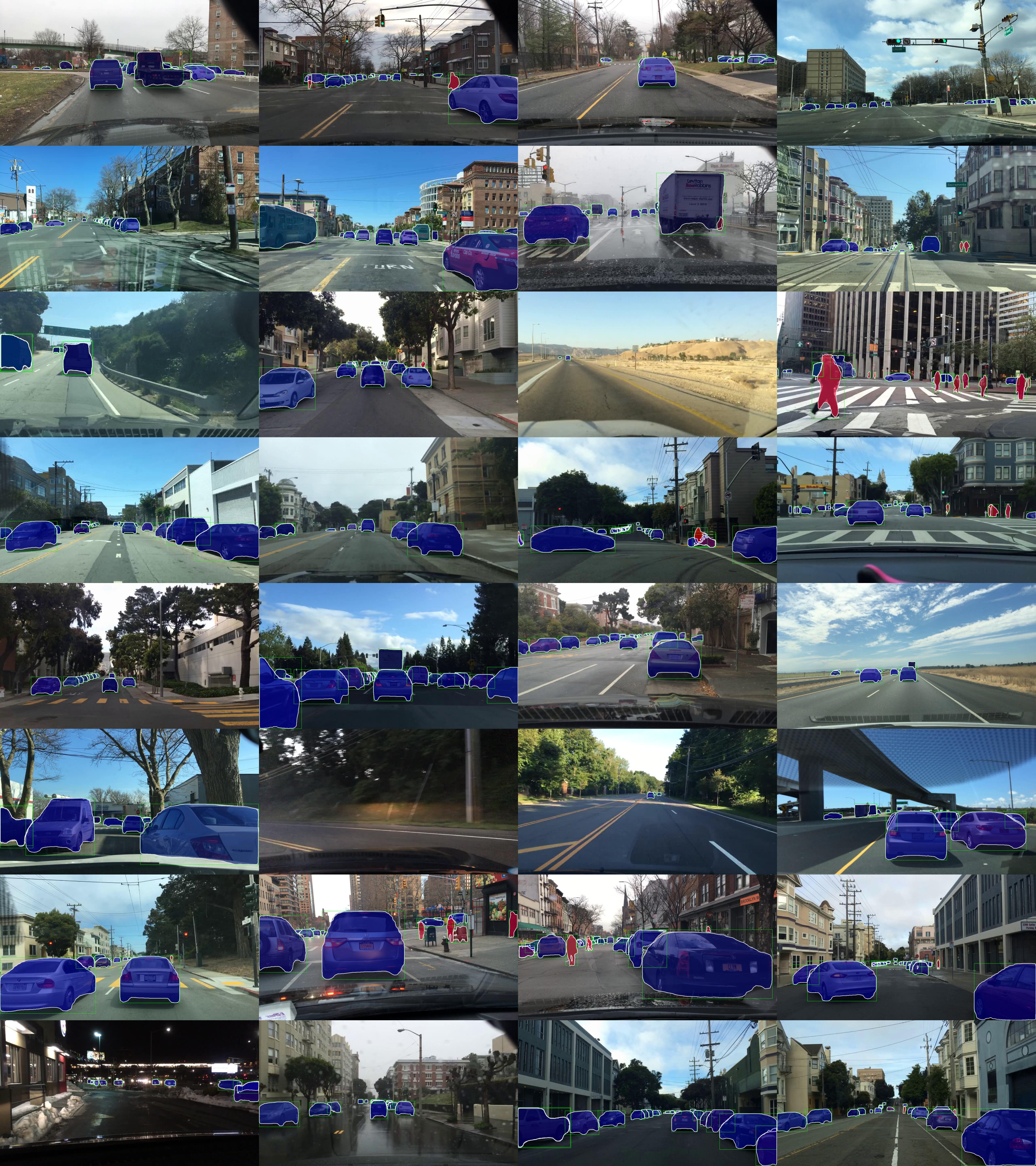}
\end{center}

\caption{\textbf{BDD100K instance segmentation visualization: pre-trained with DreamTeacher feature distillation on BDD100K.} The backbone is resnet-50, finetuned using Mask R-CNN. Only the backbone weight is pre-trained, other part of the networks are randomly initialized.
}
\label{fig:bdd100k-ins-merge-in-domain}

\end{figure*}

\begin{figure*}[t]
\begin{center}
\includegraphics[width=1.0\textwidth]{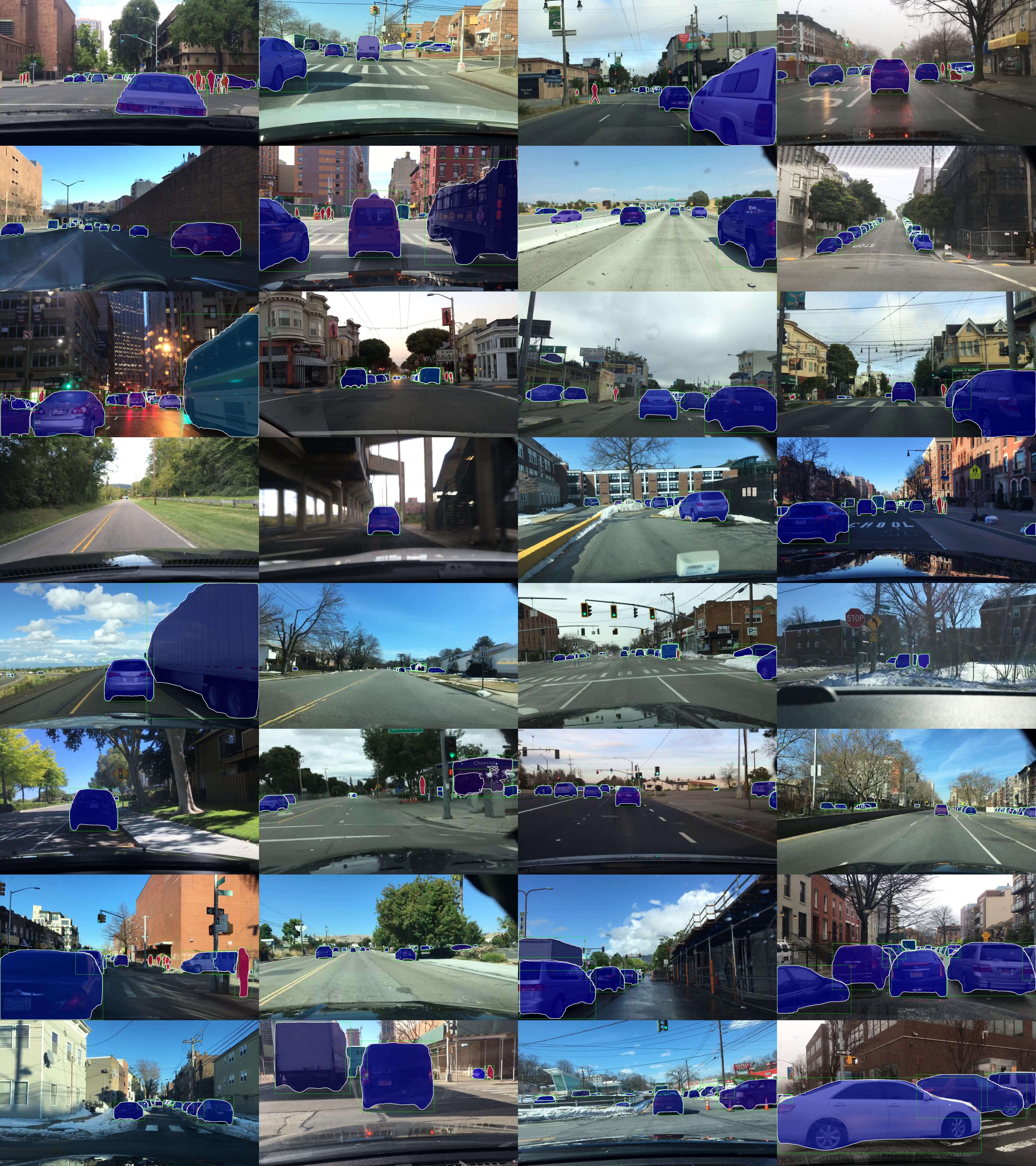}
\end{center}

\caption{\textbf{BDD100K instance segmentation visualization: pre-trained with DreamTeacher feature distillation on IN1k-1M.} The backbone is resnet-50, finetuned using Mask R-CNN. Only the backbone weight is pre-trained, other part of the networks are randomly initialized.
}
\label{fig:bdd100k-ins-merge-imagenet}

\end{figure*}

\begin{figure*}[t]
\begin{center}
\includegraphics[width=1.0\textwidth]{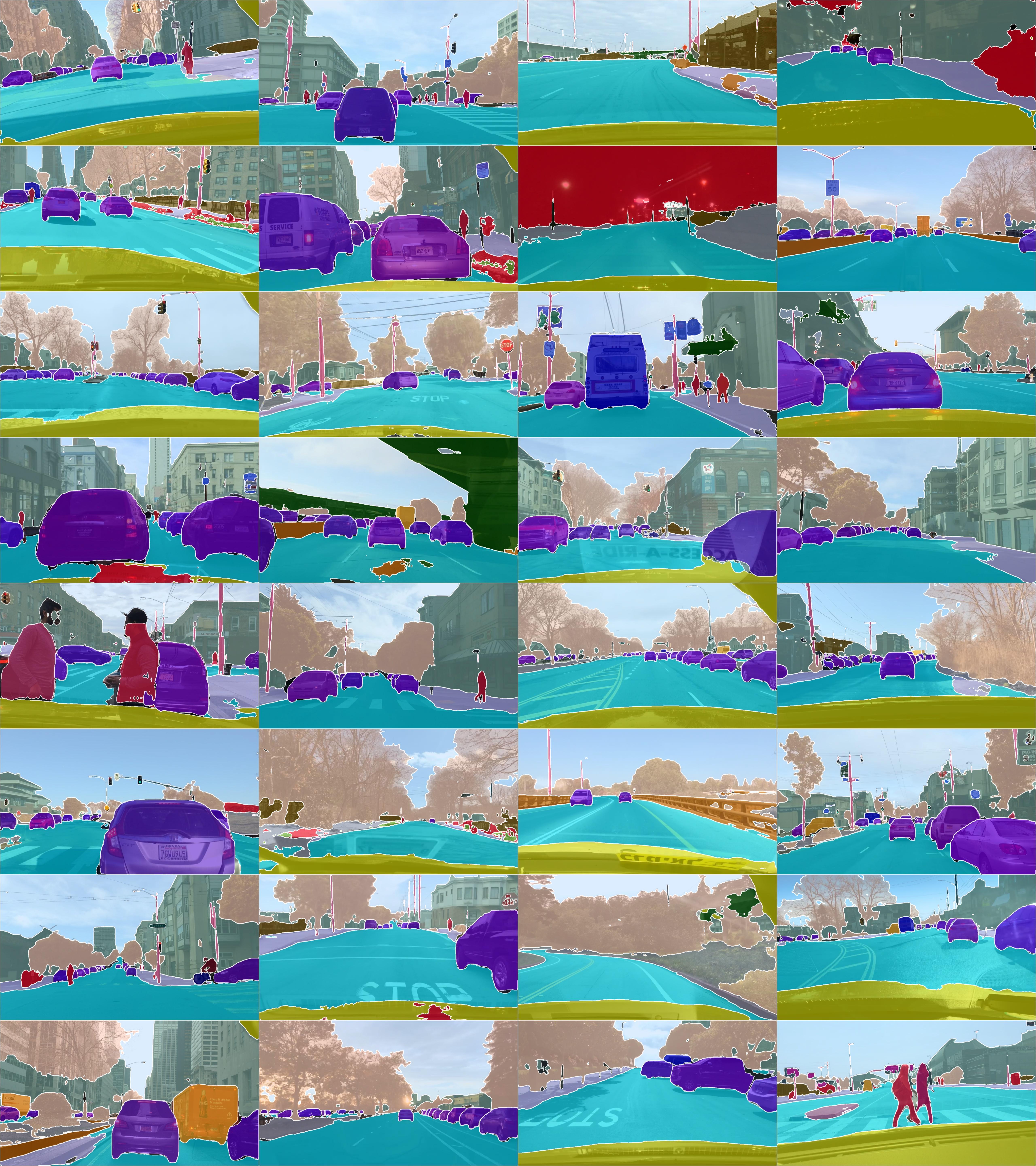}
\end{center}

\caption{\textbf{BDD100K panoptic segmentation visualization: pre-trained with DreamTeacher feature distillation on BDD100k.} The backbone is resnet-50, finetuned using PanopticFPN. Only the backbone weight is pre-trained, other part of the networks are randomly initialized.
}
\label{fig:bdd100k-pan-merge-in-domain}

\end{figure*}

\begin{figure*}[t]
\begin{center}
\includegraphics[width=1.0\textwidth]{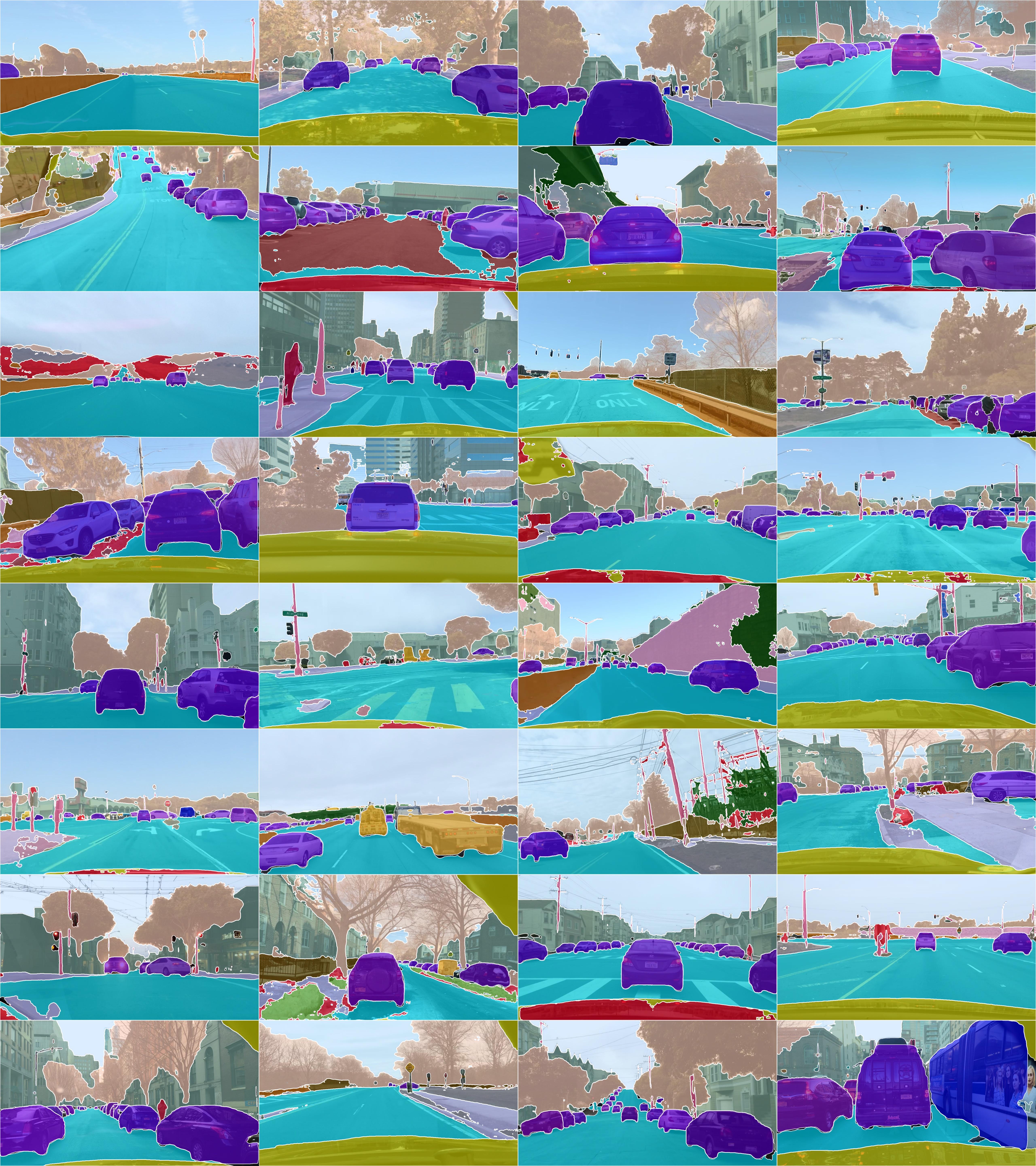}
\end{center}

\caption{\textbf{BDD100K panoptic segmentation visualization: pre-trained with DreamTeacher feature distillation on IN1k-1M.} The backbone is resnet-50, finetuned using PanopticFPN. Only the backbone weight is pre-trained, other part of the networks are randomly initialized.
}
\label{fig:bdd100k-pan-merge-imagenet}

\end{figure*}

 \fi

\end{document}